\documentclass[final,5p,times]{elsarticle}

\journal{ }
\usepackage{lineno}
\modulolinenumbers[5]

\usepackage{amsmath}
\usepackage{amssymb}
\usepackage{amsfonts}
\usepackage{bbm}
\usepackage{algorithmic}
\usepackage{textcomp}
\usepackage{xcolor}
\usepackage{multicol}
\usepackage[bookmarks=true]{hyperref}
\hypersetup{hidelinks} 
\usepackage{subcaption}
\usepackage{graphicx}
\graphicspath{{figures/}}

\usepackage[export]{adjustbox} 
\usepackage[binary-units=true]{siunitx}

\usepackage{glossaries}
\newacronym{com}{CoM}{Center of Mass}
\newacronym{DoF}{DoF}{Degrees of Freedom}
\newacronym{EoM}{EoM}{equation of motion}
\newacronym{FK}{FK}{forward kinematics}
\newacronym{IK}{IK}{inverse kinematics}
\newacronym{MPC}{MPC}{Model Predictive Control}
\newacronym{NLP}{NLP}{nonlinear programming}
\newacronym{QP}{QP}{quadratic programming}
\newacronym{TO}{TO}{trajectory optimization}
\newacronym{wrt}{w.r.t.}{with respect to}
\newacronym{FIC}{FIC}{Fractal Impedance Controller}
\newacronym{RMSE}{RMSE}{Root Mean Square Error}
\newacronym{PMP}{PMP}{Passive Motion Paradigm}
\newacronym{AICO}{AICO}{Approximate Inference Control}
\newacronym{FRI}{FRI}{Fast Research Interface}

\usepackage[ruled,vlined,linesnumbered]{algorithm2e}
\let\oldnl\nl
\newcommand{\nonl}{\renewcommand{\nl}{\let\nl\oldnl}}

\hypersetup{
    pdfauthor   = {Carlo Tiseo and Wolfgang Merkt},%
    pdftitle    = {Safe and Compliant Control of Redundant Robots Using Superimposition of Passive Task-Space Controllers},%
    pdfsubject  = {Robotics},%
    pdfkeywords = {Robots, Passive Control, Safe Control}%
}

\begin{document}

\begin{frontmatter}
    \title{Safe and Compliant Control of Redundant Robots Using Superimposition of Passive Task-Space Controllers}
    \tnotetext[title_note]{This research is supported by EPSRC UK RAI Hub ORCA (EP/R026173/1) and NCNR (EPR02572X/1), EU Horizon2020 projects MEMMO (780684) and THING (ICT-2017-1).}
    
    \author[address_edinburgh]{Carlo Tiseo\corref{corresponding_author}}
    \ead{ctiseo@ed.ac.uk}
    \author[address_edinburgh,address_oxford]{Wolfgang Merkt\corref{corresponding_author}}
    \ead{wolfgang@robots.ox.ac.uk}
    \cortext[corresponding_author]{The authors contributed equally to this work.}
    
    \author[address_edinburgh]{Wouter Wolfslag}
    \author[address_edinburgh]{Sethu Vijayakumar}
    \author[address_edinburgh]{Michael Mistry}

    \address[address_edinburgh]{Institute for Perception, Action, and Behaviour, School of Informatics, University of Edinburgh (10 Crichton St, Edinburgh, EH8 9AB, United Kingdom)}
    \address[address_oxford]{Oxford Robotics Institute, University of Oxford (23 Banbury Rd, Oxford, OX2 6NN, United Kingdom)}

    \begin{abstract}
        Safe and compliant control of dynamic systems in interaction with the environment, e.g., in shared workspaces, continues to represent a major challenge. Mismatches in the dynamic model of the robots, numerical singularities, and the intrinsic environmental unpredictability are all contributing factors.
        Online optimization of impedance controllers has recently shown great promise in addressing this challenge, however, their performance is not sufficiently robust to be deployed in challenging environments.
        This work proposes a compliant control method for redundant manipulators based on a superimposition of multiple passive task-space controllers in a hierarchy.
        Our control framework of passive controllers is inherently stable, numerically well-conditioned (as no matrix inversions are required), and computationally inexpensive (as no optimization is used).
        We leverage and introduce a novel stiffness profile for a recently proposed passive controller with smooth transitions between the divergence and convergence phases making it particularly suitable when multiple passive controllers are combined through superimposition.
        Our experimental results demonstrate that the proposed method achieves sub-centimeter tracking performance during demanding dynamic tasks with fast-changing references, while remaining safe to interact with and robust to singularities. 
        The proposed framework achieves such results without knowledge of the robot dynamics and thanks to its passivity is intrinsically stable.
        The data further show that the robot can fully take advantage of the redundancy to maintain the primary task accuracy while compensating for unknown environmental interactions, which is not possible from current frameworks that require accurate contact information.
    \end{abstract}


\end{frontmatter}

\section{Introduction}
Redundant robots---which have more \gls{DoF} than required for a task---have been widely studied and deployed due to their intrinsic flexibility. 
The higher dimensionality of the joint configuration space \gls{wrt} the task space makes these systems more adaptable as multiple solutions can be found. 
\begin{figure}[t]
    \centering
    \includegraphics[width=0.8\linewidth,trim={0cm 5cm 0cm 1cm},clip]{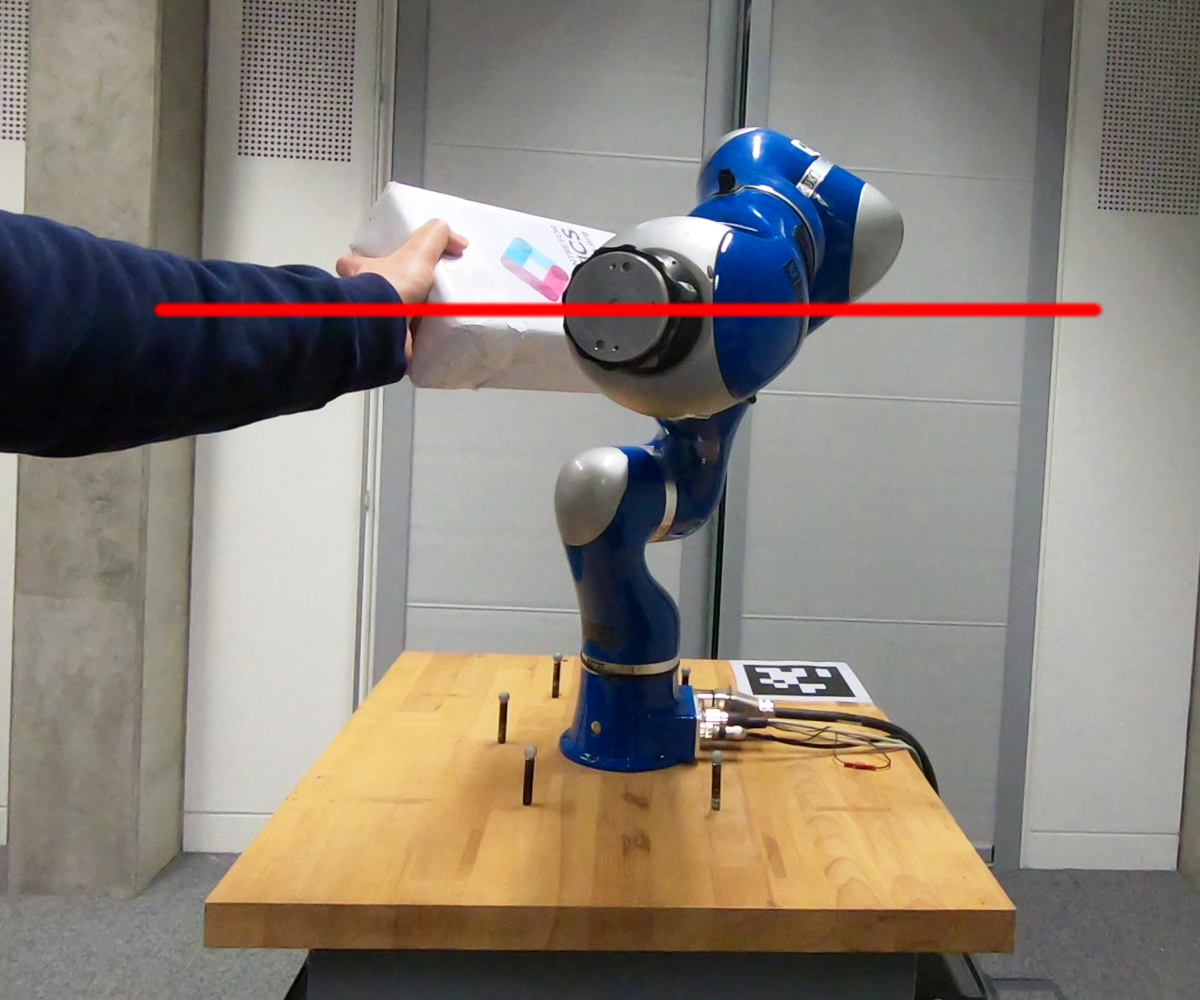}
    \caption{Stack of Passive Controllers executing a reference motion to follow a line trajectory while an unknown external disturbance is introduced to the elbow joint. The controller adapts safely and compliantly to the disturbance and degrades tracking performance gracefully.} \label{fig:intro}
\end{figure}
However, this flexibility introduces a higher complexity for both planning and control that rapidly increases with the system and task dimensionality.
For example, computing the joint configuration from a task-space pose, i.e. \gls{IK}, becomes increasingly more challenging with the increase of the redundancy dimension \cite{siciliano,d2001learning}. 
This problem is also encountered when dealing with the inverse dynamics problem, which is used to derive the control laws used in interaction control \cite{siciliano,xin2020,keppler2018}.
To address these inverse problems in rigid systems, multiple optimization frameworks and approaches to deal with challenges arising from numerical conditioning have been developed.
Currently, inverse problems for soft robots\footnote{Here, we use soft robots to refer both to systems made from non-rigid materials as well as those with compliant control, e.g., collaborative robots.} and optimization of the task-space dynamics are still open problems \cite{bruder2019}.
A recent approach to robustness for achieving task-space compliance behaviors include systems for increasing the robustness of projections through software by modulating/adapting the references \cite{xin2020}. This robustness can also be achieved by exploiting more complex hardware design that embeds variable mechanical compliance directly into the robot structure and its actuation \cite{keppler2018,braun2013robots}. As an example, \textit{Keppler et al.} have proposed a control architecture that enables to retain both robustness and accurate tracking \cite{keppler2018} while related work has algorithmically optimised spatiotemporal modulation of impedance to achieve tasks more efficientl \cite{nakanishi2011stiffness}. These approaches takes advantages of hardware equipped with Variable Stiffness Actuators (VSAs) --which allows better dynamic performances from the hardware, but they also increase the complexity and cost of motion planning with these system besides the need for very accurate modelling of the VSA structures.

\begin{figure*}[t]
    \centering 
    \includegraphics[width=0.8\linewidth]{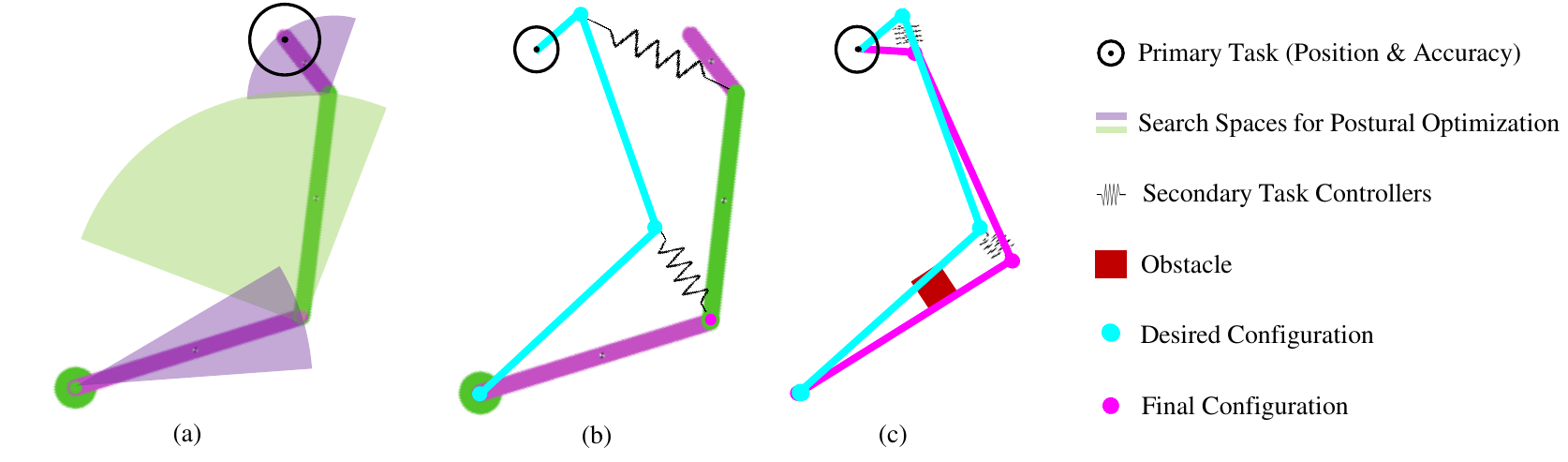}
    \caption{Conceptual overview of the stack of passive task-space controllers framework:
    (a) The primary task is defined in terms of desired position, accuracy, and maximum exerted force produces a non-linear impedance profile to constrain the robot's end-effector ($3^{\text{rd}}$-link).
    Impedance profiles acting on additional links can subsequently be defined to control the robot joint-space configuration. Thus, the redundant manipulator is controlled via the interaction of several task-space controllers.
    (b) Exemplification of how the secondary tasks controllers pull the robot towards the desired configuration
    (c) In case an unknown obstacle is found on one of the links' paths, the primary task controller guarantees that the end-effector gets as close as possible to the desired end-effector position despite the joint configuration converges to a different solution. Note, this adaptation is performed at the control level and does not involve replanning.}
    \label{fig:conceptual_overview}
\end{figure*}
Inverse kinematics solutions and task-space dynamics projections are required for controlling redundant robots and they share similar challenges, as analyzed  in depth in \cite{dietrich2015,khatib1993,Vijayakumar-RSS-19}. 
In summary, both problems rely on the inversion of the Jacobian matrix, which is non-square in redundant manipulator due to the different task and joint space dimensions \cite{siciliano}.
The pseudo-inverse is a transformation that solves such a problem. 
It separates the information regarding robot states into two orthogonal sub-spaces (task-space and null-space), which are not expected to exchange information (i.e., energy). 
Therefore, retaining the orthogonality between these two sub-spaces is paramount for the algorithms' stability \cite{dietrich2015,khatib1993,Vijayakumar-RSS-19}.
Maintaining this orthogonality depends on both the robot kinematics and the task -- which can be quite difficult to achieve and maintain, especially during highly variable situations, such as sudden changes in contacts and dynamic interactions (\autoref{fig:intro}). 
As one example, the dynamically consistent inverse obtains orthogonality via the minimization of the kinetic energy projected by the null-space into the task-space \cite{Vijayakumar-RSS-19, khatib1987unified}.

Passive controllers have been proposed to theoretically guarantee interaction stability under uncertain interaction conditions, using for instance virtual tanks as energy storage (i.e., path integral) for the non-conservative energy of the controller. 
However, their passive behavior trades-off tracking performances to retain safety of interaction, making this framework difficult to deploy in highly variable environments \cite{Dietrich2016}. The result presented in their manuscript focuses on verifying passivity and safety of interaction and does not provide a clear quantitative analysis of task-space tracking performance. However, based on the figure of the Cartesian tracking error reported for the simulation results for a 4-\gls{DoF} planar manipulator, it seems that we can expect about \SI{1}{\centi\meter} residual pose error in the best case scenario (i.e., there is energy in tank).
Moreover, virtual tank impedance controllers are only passive if there is energy left in their virtual tanks.
Due to passivity constraints, the tanks' energy can be only charged from external energy sources \cite{Dietrich2016}.
Realising passive control is made even more difficult when dealing with null-space and task-space controllers.
In fact, as they are orthogonal to each other, tracking the total energy exchanged by the manipulator is challenging \cite{babarahmati2019}. 
Another challenge to stability of virtual tank controllers is to maintain the orthogonality between null-space and task-spaces during highly variable tasks. 
Higher non-linearity in the dynamics reduces the accuracy in the computation of the orthogonal projection that, consequentially, generates unaccounted energy transfer between the two sub-spaces \cite{Vijayakumar-RSS-19,babarahmati2019}.

This work investigates the possibility of using superimposition of passive task-space controllers to drive redundant manipulators rather than relying on null-space controllers, cf. \autoref{fig:conceptual_overview}. 
Conceptually, this follows the idea to generate task-space wrenches at multiple links and map them back to joint-level torques.
To do so, the proposed solution will not need to rely on any mathematical projections (and implicitly, matrix inversions) required by the null-space projections.
Virtual mechanical constraints are instead generated using a superimposition of task-space controllers to control task-space and the redundant degrees of freedom of the manipulators. 
However, implementing such a solution will require a controller framework that is intrinsically stable. 
The recently proposed \gls{FIC} \cite{babarahmati2019} is a passive controller meeting this requirement.
It relies upon a non-linear stiffness behavior in the task-space to track the energy exchanged between the robot and the environment, and treats the unexpected energy flow from the null-space as an external perturbation. 
The controller uses the concept of fractal impedance for the implementation of a passive controller that can provide good performances in both trajectory and force tracking.  
Thereby, it detaches the robot stability from the postural optimization, which are currently bounded for interaction controllers relying on \gls{QP} optimization \cite{xin2020}. Furthermore, our proposed method is independent of any specific type of actuation and thus can be used in any torque/force controlled robot.

In summary, our contributions are:
\begin{enumerate}
    \item \emph{Superimposition of Passive Task-Space Controllers} to preserve the primary task by sacrificing a secondary tasks through exploitation of the mechanical redundancy. The priority of the controller is determined by the maximum force exertable by the controller, as will be explained in \autoref{sec:method}. As the framework only relies on the forward computation of kinematics and Jacobian, it is numerically stable and computationally inexpensive. Further, it can be used with uncertain and imprecise dynamics models as it only relies on the kinematics model (\ref{sec:StackOfFIC}).
    \item Proposal of a new force profile for a Fractal Attractor which enables a smooth transition between convergence and divergence phases (\ref{sec:SigFractImp}).
    \item Validation of our approach both in simulation to test contact interaction with unknown obstacles and in real hardware experiments (\ref{sec:experiments}) to evaluate reference tracking performance for fast reference motion (\ref{sec:results}).
    As all open parameters have a physically tractable meaning and the controller is intrinsically stable, online tuning can safely be performed.
\end{enumerate}

We intend to open source our implementation for simulation and hardware experiments with the publication of this manuscript.

\section{Method} \label{sec:method}
The null-space of a redundant manipulator is a set of joint-space configurations having the same end-effector pose. Therefore, null-space optimization frameworks identify the optimal joint-space configuration for a given task. Stack of Task optimization methods are iterative algorithms applying Null-Space optimization to a hierarchy of tasks. Their main limitation is that null-space projections are inserted in the control loop, rendering the controllers susceptible to numerical instability connected with the null-space projections. 

The proposed method, shown in \autoref{fig:conceptual_overview}, aims to remove null-space projections from the control loop. Null-space projections are used to account for the external interaction in the whole-body control optimization problem \cite{xin2020}. This type of formulation requires not only to make \textit{a priori} assumptions on the environmental interaction, but it also renders the controller stability dependent on their accuracy. Thus, the controller stability is highly susceptible to erroneous assumptions, which lead only to sub-optimal behaviour in the best case scenario \cite{xin2020}.

The proposed method unravels the co-dependency between stability and assumptions made on the external environment by using a superimposition of task-space controllers that generates virtual force field (i.e., soft mechanical constraints) that pull the robot towards the desired configuration. This is a different approach to handling redundancy compared to the null-space approach. The superimposition of the controller will bias the robot to move towards a certain preferred posture, without guaranteeing that this particular configuration will be reached. In fact, the controller will continuously maintain a mechanical equilibrium between the virtual forces and the environmental interaction without requiring any assumption on the environmental interaction. This implies that the controller is robust to unknown environmental interaction, but is not guaranteed to be in a global optimum. However, it is likely to settle in the closest minimum in the system energetic manifold (i.e., the closest state with mechanical equilibrium).

In synthesis, the relative strength of these virtual constraints will determine the trade-off between the tasks assigned to the controllers and, consequentially, the order in which the task accuracy will be sacrificed. While this method can be applied with any type of task-space controllers, using passive controllers guarantees stability by independently verifying that all the superimposed controllers are stable. Among the different passive controllers, we have chosen the Fractal Impedance Controller due to its explicit formulation of the tasks in terms of virtual mechanical constraints (i.e., desired force/displacement behavior), enabling direct control of the controllers' trade-off policies. For the scope of this paper, an optimization-based inverse kinematics algorithm was used to obtain the reference configuration (\emph{postural optimization}). Forward kinematics is subsequently used to extract the task-space references for the individual superimposed controllers. In place of this postural optimization, more comprehensive planners and frameworks could be used to provide and update the reference configurations.

\subsection{Inverse Problem and Kineto-Static Duality}
\label{sec:IPM}
The generalized inverse of $A \in \mathbb{R}^{n \times m}$ is defined as any matrix $G\in \mathbb{R}^{m \times n}$ that satisfies the following equations:

\begin{equation}
    \label{Ginverse}
     \begin{cases}
     \vec{a}=G \vec{b}+ (I_{n}-GA)\vec{a}_\epsilon=G \vec{b}+ P \vec{a}_\epsilon\\
     AGA-A=0
     \end{cases}
\end{equation}
where  $\vec{a} \in \mathbb{R}^{n}$, $\vec{b} \in \mathbb{R}^{m}$, $\vec{a}_\epsilon \in \mathbb{R}^{n}$ and $I_n \in \mathbb{R}^{n \times n}$ is the identity matrix. $P$ is a projection matrix that projects a generic vector $a_\epsilon$ into the null-space of $A$, $\mathcal{N}(A)$. 
Redundant robots are more flexible than non-redundant systems, however, they do not have a bijective transformation between generalized coordinates and task-space. 
Thus, control algorithms rely on numerical optimization to solve the inverse problem and identify viable strategies. This is task dependent and degenerates when $A$ drops rank (i.e., $det(A)=0$) \cite{Vijayakumar-RSS-19,xin2020}. 
Specifically, the rank of the inverse projection matrix drops if the robot is in a singular configuration or the task constraints are violated (e.g., unexpected sudden loss of contact) \cite{Vijayakumar-RSS-19}. 

The idea of taking advantage of the kineto-static duality to address the inverse problem has been introduced with the concept of Port-Hamiltonian control in \cite{hogan1985impedanceP1,hogan1985impedanceP2}. In fact, the kinematic joint-space information can be used to derive task-space behavior and task-space force interaction can be used to relate back to joint-space torques:
\begin{equation}
    \label{kinetostatic}
    \begin{cases}
     \vec{\nu}=J \dot{\vec{q}} \\
     \vec{\tau}=J^\text{T} \vec{h}
    \end{cases}
\end{equation}
where $J\in \mathbb{R}^{n \times m}$ is the geometric Jacobian matrix, $\vec{\nu} \in \mathbb{R}^n$ is the end-effector twist, $\vec{\dot{q}} \in \mathbb{R}^m$ is the joint velocities' vector and $\vec{h} \in \mathbb{R}^n$ is the end-effector wrench, $\vec{\tau} \in \mathbb{R}^m$ is the joint torques' vector. 

\subsection{Fractal Impedance Controller} \label{sec:FractImp}
\begin{figure}[t]
    \centering
    \begin{subfigure}[b]{0.8\linewidth}
        \centering 
        \includegraphics[width=\linewidth]{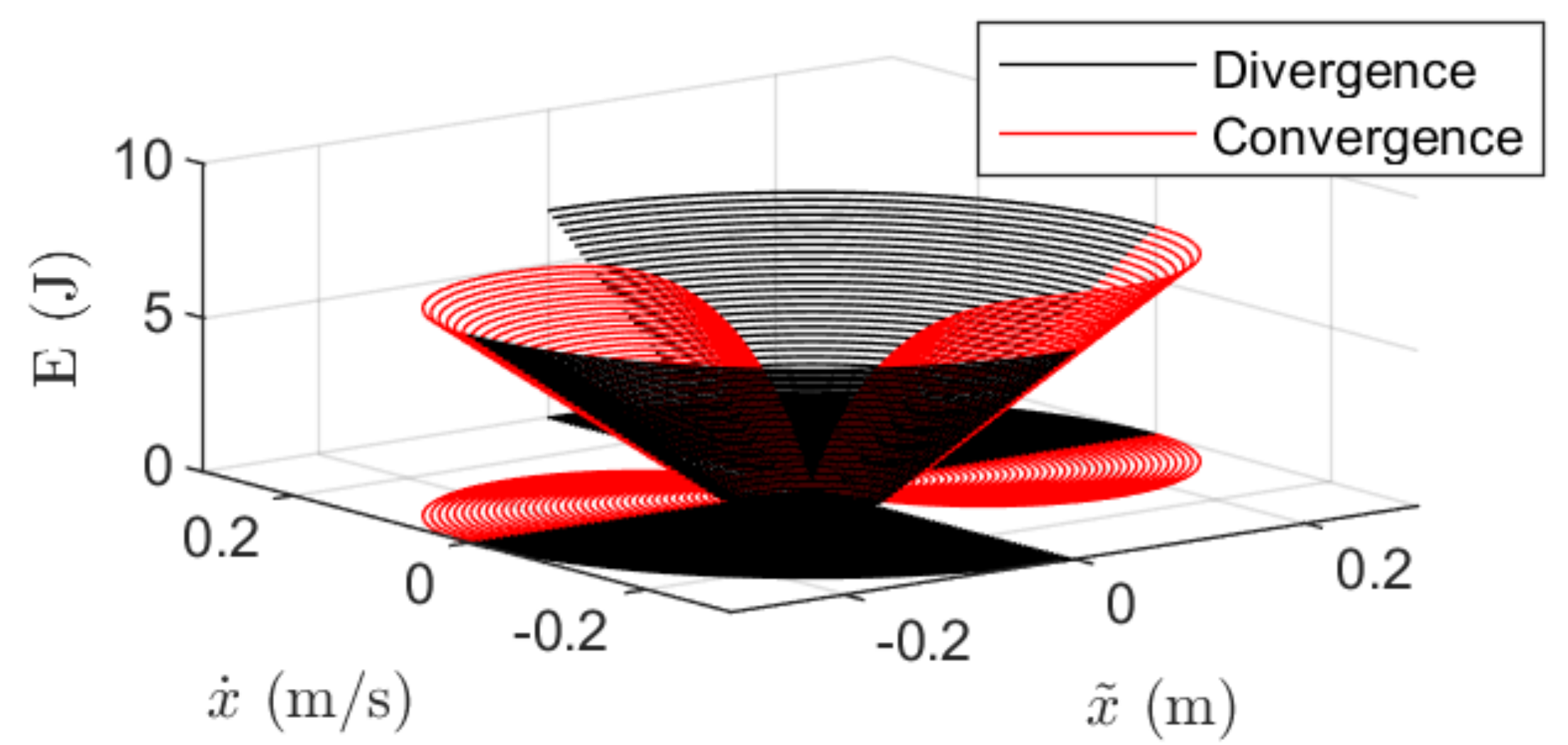}
        \caption{}\label{fig:2b}
    \end{subfigure}\\
    \begin{subfigure}[b]{0.8\linewidth}
        \centering 
        \includegraphics[width=\linewidth]{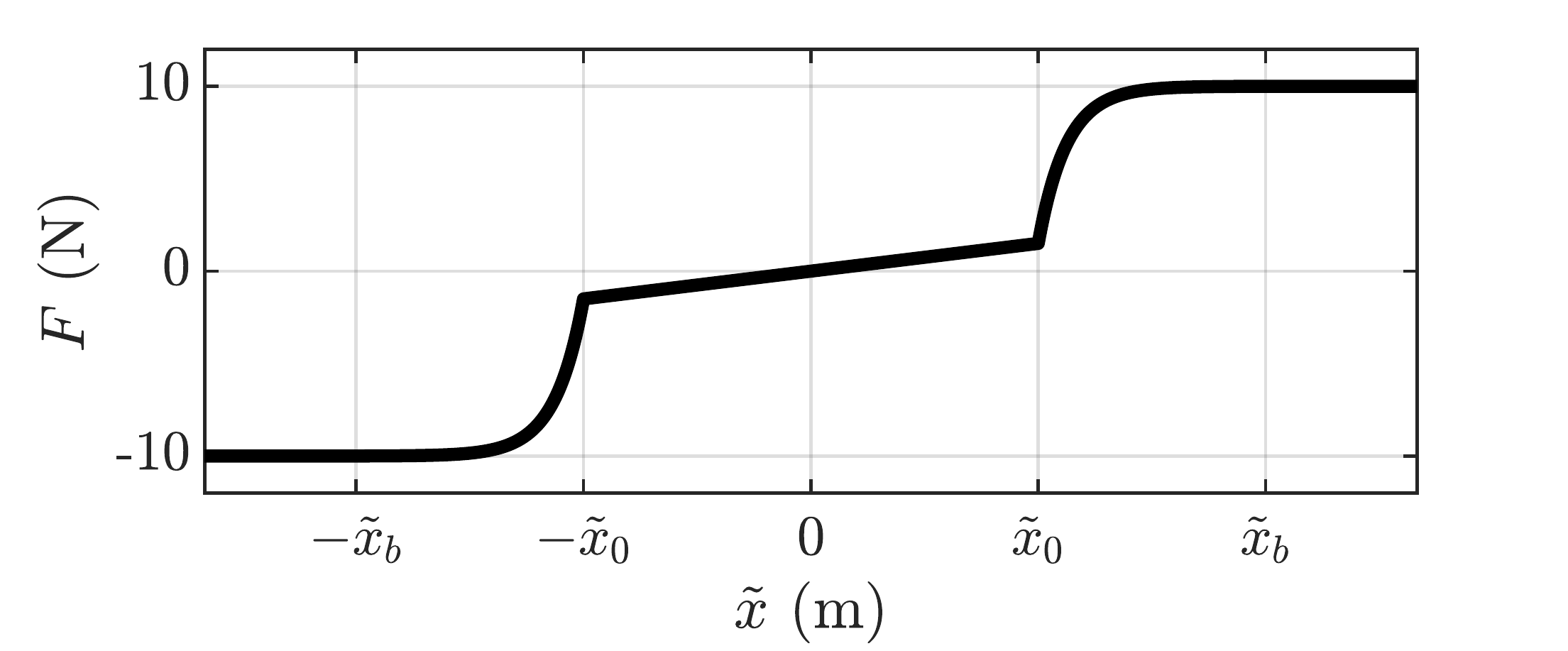}
        \caption{}\label{fig:2c}
    \end{subfigure}\\
    \begin{subfigure}[b]{0.8\linewidth}
        \centering 
        \includegraphics[width=\linewidth]{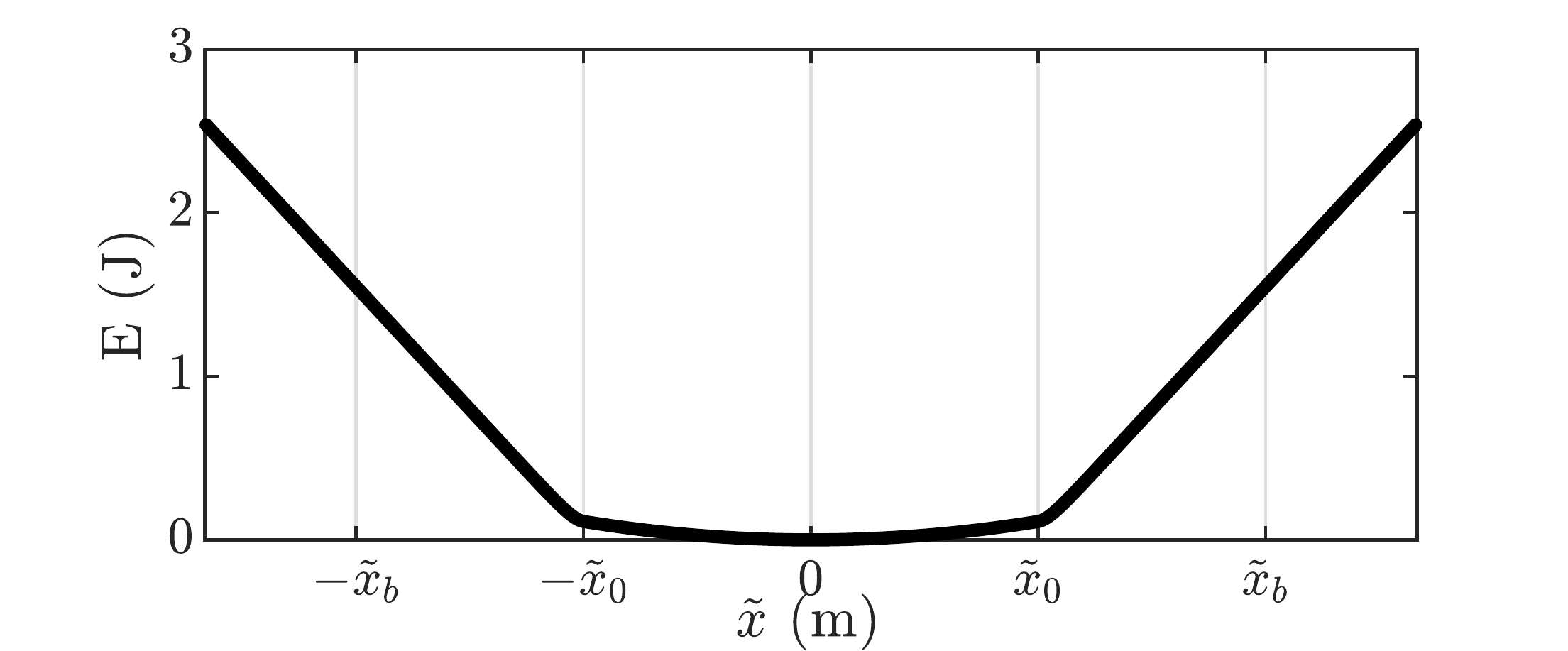}
        \caption{}\label{fig:2d}
    \end{subfigure}
    \caption{
    (a) \textbf{Attractor's Phase Portrait} \textemdash ~Fractal impedance Controller using the proposed spring (\ref{sec:SigFractImp}), which differently from the linear spring has a compliant region near the fixed point in the desired state. The trajectories beyond $\tilde{x}>0.3$ have been omitted 
    (b) Proposed force profile. 
    (c) Energy profile associated with the proposed force profile.
    }
    \label{fig:fractal_attractor}
\end{figure}
The FIC controls the robot as a non-linear mass-spring system, and generates the attractor in \autoref{fig:2b} around the desired state. The equivalent mechanical system equation is:
\begin{equation*}
    \Lambda_c (\vec{q}) \vec{\Ddot{x}} + n(q,\dot{q}) + K(\vec{\tilde{x}}) \vec{\tilde{x}}=F_{Ext}
\end{equation*}
\noindent where $\Lambda_c (\vec{q})$ is the projection of the task-space inertia matrix at the end-effector, $n(q,\dot{q})$ the non-linear robot dynamics and $F_{Ext}$ is the external force.
The state-dependent stiffness gain $K(\vec{\tilde{x}})$ is derived from the desired end-effector interaction properties (i.e., force/displacement), which can be regulated online without affecting stability \cite{babarahmati2019}. For completeness, we provide a proof of stability in \ref{sec:stabilityAnalysis}.
The attractor is implemented using a switching behavior that introduces an additional nonlinear spring which triggers when the system starts converging (i.e., zero crossing of $\vec{\dot{x}}$). 
The updated impedance conserves the energy accumulated in the controller while diverging and redistributes the energy altering the trajectory during the convergence, as shown in \autoref{fig:fractal_attractor}.
Therefore, the stability of the controller is guaranteed by the fractal attractor (\autoref{fig:fractal_attractor}). This determines the passivity of the controller and the online adaptability; it is independent of the chosen impedance. 

For each \gls{DoF} in the task-space, the \gls{FIC} is given in \autoref{alg:FIC}. The control torques ($\vec{\tau}_{\text{ctr}}$) can be calculated from $\vec{h}_e \in \mathbb{R}^6$ using \eqref{kinetostatic}. Differently from the \gls{FIC} control scheme introduced in \cite{babarahmati2019} on a sharp force/torque saturation, this manuscript introduces a more flexible force profile. The new force profile allows to independently tune the linear, non-linear and saturation behaviors of the controller wrench, making it easier to tune the controller for different tasks.

\begin{algorithm} 
\SetAlgoLined
\SetKwData{Left}{left}
\SetKwData{Up}{up}
\SetKwFunction{FindCompress}{FindCompress}
\SetKwInOut{Input}{input}
\SetKwInOut{Output}{output}
\Input{Convergence$/$Divergence, $\tilde{x}$, $\tilde{x}_{\text{max}}$}
\Output{$h_{\text{e}}$}
  \eIf{diverging from ${x_{\text{d}}}$}{
     $h_{\text{e}}=f(\tilde{x})= K(\tilde{x})\tilde{x}$\\
  }{
     $h_{\text{e}} = \frac{4 E(\tilde{x}_{\text{max}})}{\tilde{x}_{\text{max}}^2} (0.5\tilde{x}_{\text{max}} -\tilde{x})$\\
  }
 \nonl where:\\
 \nonl ${x_{\text{d}}}$ is the desired position \\
 \nonl $h_{\text{e}}$ is the desired force at the end-effector \\
 \nonl $\tilde{x}$ is the pose error \\
 \nonl $K(\tilde{x})$ is the nonlinear stiffness\\
 \nonl $x_\text{e}$ is the end-effector position\\
 \nonl $\tilde{x}=x_\text{d}-x_\text{e}$ is the position error \\
 \nonl $\tilde{x}_{\text{max}}$ is the maximum displacement reached at the end of the divergence phase \\
 \nonl $E$ is the energy associated with the divergence profile of the impedance controller \\
 \caption{Mono-dimensional \gls{FIC}}
 \label{alg:FIC}
\end{algorithm}

\subsection{Sigmoidal Force Profile for Fractal Impedance} \label{sec:SigFractImp}
The \gls{FIC} relies on a stiffness profile. 
The profile proposed in \cite{babarahmati2019} results in fast changes in stiffness, and only allows limited task-dependent tuning of the profile. 
Therefore, we propose a sigmoidal force profile for an easier definition of the stiffness profile, allowing to better adapt the robot impedance behavior to the different task.
Similarly to the profile proposed by \cite{babarahmati2019}, the sigmoidal profile is fully determined based on the maximum force ($F_{\text{Max}}$) to be exerted at a chosen position error ($|\tilde{x}|=\tilde{x}_{\text{b}}$).
Here, the position error is defined as the difference between the desired end-effector pose and the current pose ($\tilde{x}=x_d-x$).
We introduce an additional displacement parameter ($|\tilde{x}|=\tilde{x}_{0}$) which describes the minimum displacement to activate the nonlinear impedance, as shown in \autoref{fig:2c}.
The proposed force profile thus becomes:
\begin{equation}
    \label{ForceProf}
    F_{\text{K}}=\left\{
    \begin{array}{ll}
      K_0 \tilde{x},  & |\tilde{x}|<\tilde{x}_{0}\\\\
     \text{sgn} (\tilde{x}) (\Delta F (1-e^{-\frac{|\tilde{x}| - \tilde{x}_0}{b}})+& \\
     + K_0 \tilde{x}_0),   & \tilde{x}_{0}\le |\tilde{x}|<\tilde{x}_{b}\\\\
     \text{sgn}(\tilde{x})  F_{\text{Max}},   & \text{Otherwise}
    \end{array}\right.
\end{equation}
where $b=(\tilde{x}_\text{b}-\tilde{x}_\text{0})/S$ is the characteristic length, $S$ determines the shape of the sigmoid curve and $\Delta F =(F_{\text{Max}} - K_0.\tilde{x}_0)$. In this work, we use $S=20$ to ensure force saturation before $\tilde{x}_\text{b}$. 

The proposed force profile can further be associated to an energy (\autoref{fig:2d}) that is an unbounded Lipschitz function. 
It therefore respects the requirement for Lyapunov's stability by the fractal attractor controller \cite{babarahmati2019}.
For the proposed force profile, this becomes:
\begin{equation}
    \label{EnergyProf}
    E_{\text{K}}=\left\{
    \begin{array}{ll}
      0.5 K_0 \tilde{x}^2,  & |\tilde{x}|<\tilde{x}_{0}\\\\
      F_{\text{Max}}|\tilde{x}| -&\\
      +F_{\text{Max}}\tilde{x}_0 + (K_0\tilde{x}_0^2)/2 -&\\
      +(1-e^{-\frac{|\tilde{x}| - \tilde{x}_0}{b}})b\Delta F, 
      & \tilde{x}_{0}\le |\tilde{x}|<\tilde{x}_{b}\\\\
      F_{\text{Max}}|\tilde{x}|-&\\
      +F_{\text{Max}}\tilde{x}_0 + (K_0\tilde{x}_0^2)/2 -&\\
      +(1-e^{-\frac{\tilde{x}_b - \tilde{x}_0}{b}})b\Delta F,   & \text{Otherwise}
    \end{array}\right.
\end{equation}

\subsection{Controller Superimposition for the Control of Redundant Robots} \label{sec:StackOfFIC}
We propose to implement the same solution using virtual soft mechanical constraints generated by a superimposition of task-space controllers that drive the robot to assume a commanded reference posture. 
The benefit of using impedance controllers based on fractal impedance is that their passivity allows for superimposition without compromising overall system stability. 
Therefore, the total torque vector ($\vec{\tau_{\text{tot}}}$) can be computed by the superimposition of controllers as:
\begin{equation}
    \label{TControllerStack}
    \begin{array}{ll}
        \vec{\tau}_{\text{tot}}=\sum_{i=1}^n J_i^\text{T}h_{ei}
    \end{array}
\end{equation}
where $J_i$ and $h_{ei}$ are the Jacobian and the wrench generated by the impedance controller of the $i^{th}$-link, as depicted in \autoref{fig:conceptual_overview}.

\section{Evaluation} \label{sec:experiments}
We evaluate our proposed method using a 7-\gls{DoF} torque-controlled Kuka LWR3+ manipulator in both simulation and hardware experiments. We apply a superimposition of two task-space controllers: A 6-\gls{DoF} \gls{FIC} controller at the end-effector ($7^{th}$ link) and a 3-\gls{DoF} \gls{FIC} controller at the elbow ($4^{th}$ link of the KuKA URDF) for postural control.
\begin{figure*}[t]
   \vspace{-6pt}
   \includegraphics[width=0.495\linewidth,trim={8cm 8cm 0cm 16cm},clip]{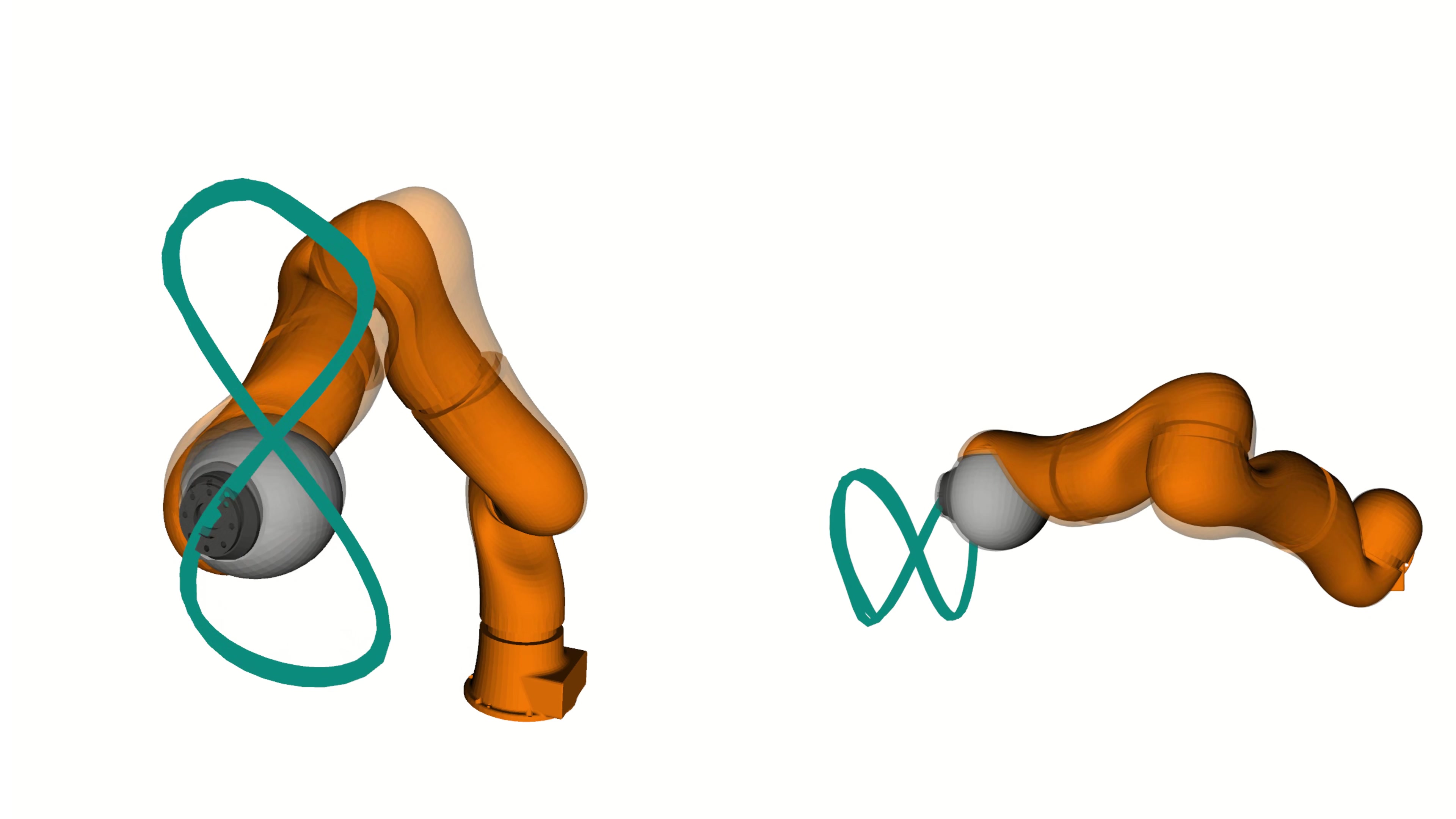}
   \hfill
   \includegraphics[width=0.495\linewidth,trim={8cm 8cm 0cm 16cm},clip]{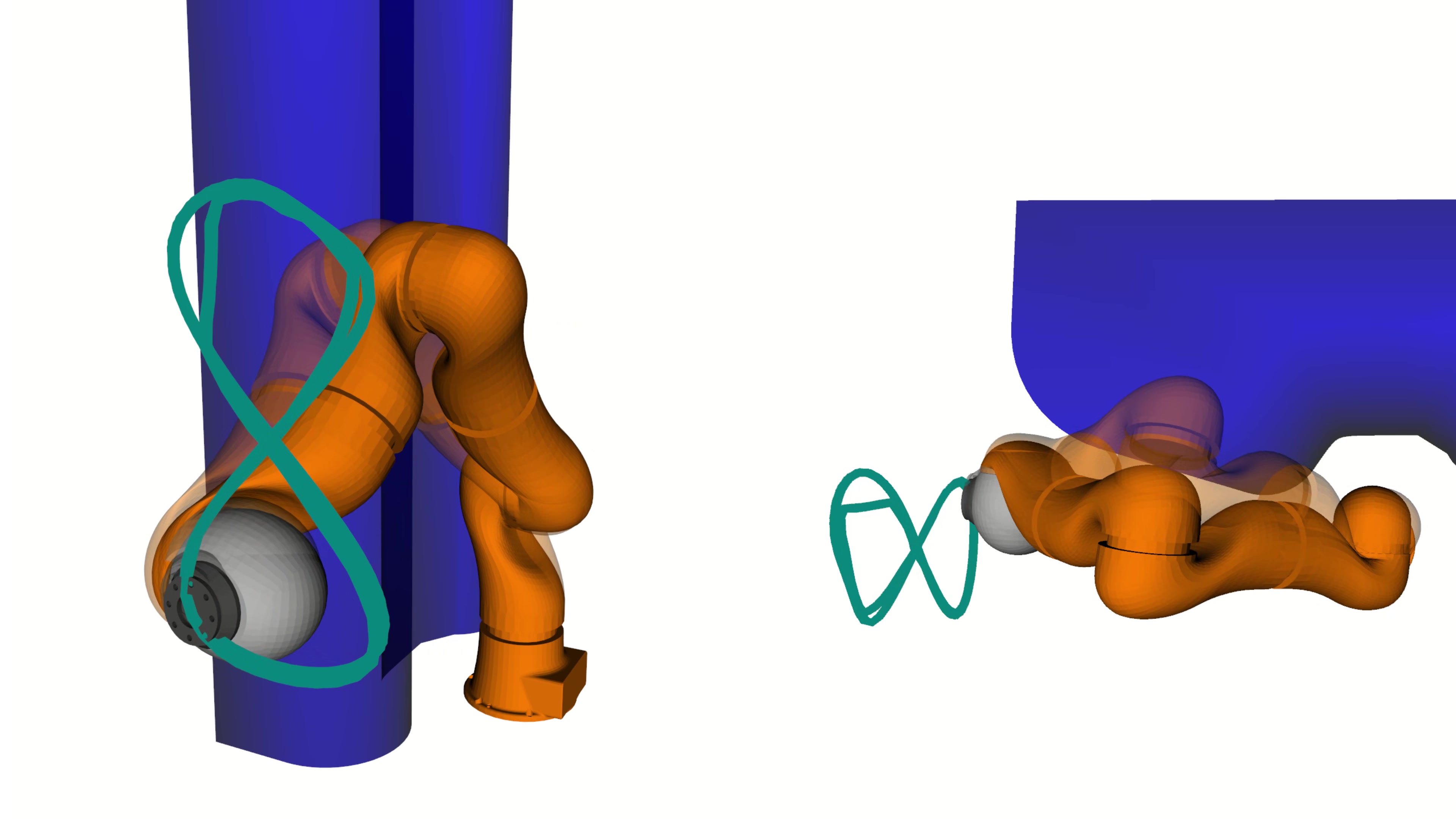}
   \caption{
       Interaction with an unsensed object in simulation.
       The planned reference motion is shown with an alpha value while the real robot state is shown in solid, with the trace of the end-effector for both overlayed.
       Note, the controller remains in contact and follows the surface of the obstacle.}
   \label{fig:gazebo_unsensed_force_interaction}
\end{figure*}

\begin{table*}[!htb]
\centering
\begin{tabular}{r|r|r|r|r|r|r|r|r|r|}
\cline{2-10}
                                       & \multicolumn{9}{c|}{\textbf{Simulations}}                                                                                                                                                                           \\ \cline{2-10} 
                                       & \multicolumn{3}{c|}{Elbow Controller}                                                                                                                                                                                                      & \multicolumn{6}{c|}{End-Effector Controller}                                                                                                                                                     \\ \hline
                                       \multicolumn{1}{|l|}{$x_0$ (\si{\meter} or \si{\radian})} & $.01$ & $.01$ & $.01$  &  $.005$  & $.005$  & $.005$  &  $.057$ & $.057$ & $.057$  \\ \hline
\multicolumn{1}{|l|}{$x_\text{b}$ (\si{\meter} or \si{\radian})}& $.011$  & $.011$  & $.011$  & $.006$  & $.006$ & $.006$ & $.63$ & $.63$& $.63$ \\ \hline
\multicolumn{1}{|l|}{$F_{\text{Max}}$ (\si{\newton} or \si{\newton\meter})} &$50$&$50$&$50$&$150$&$150$&$150$&$100$&$100$&$100$\\ \hline
\multicolumn{1}{|l|}{$K_0$(\si{\newton\per\meter} or \si{\newton\meter\per\radian})} &  $5000$& $5000$  & $5000$ &$5000$&$5000$&$5000$&$100$&$100$&$100$\\\hline

\multicolumn{1}{c|}{}                  & \multicolumn{9}{c|}{\textbf{Robot}}                                                                                                                                                                                                                                                                                                                                                                                                                   \\ \cline{2-10} 
                                       & \multicolumn{3}{c|}{Elbow Controller}                                                                                                                                                                                                      & \multicolumn{6}{c|}{End-Effector Controller}                                                                                                                                                     \\ \hline
                                       \multicolumn{1}{|l|}{$x_0$ (\si{\meter} or \si{\radian})} & $.01$ & $.01$ & $.01$  & $.06$  & $.06$  & $.06$ &NA&NA&NA  \\ \hline
\multicolumn{1}{|l|}{$x_\text{b}$ (\si{\meter} or \si{\radian})}& $.011$  & $.011$  & $.011$ & $.08$  & $.08$ & $.08$ &NA&NA&NA \\ \hline
\multicolumn{1}{|l|}{$F_{\text{Max}}$ (\si{\newton} or \si{\newton\meter})} &$10$&$10$&$10$&$100$&$100$&$100$&NA&NA&NA\\ \hline
\multicolumn{1}{|l|}{$K_0$ (\si{\newton\per\meter} or \si{\newton\meter\per\radian})} & $630$ & $630$ & $630$ &$800$&$800$&$800$&NA&NA&NA\\ \hline

\end{tabular}
\caption{\gls{FIC} Controller Parameters used for simulation and hardware experiments.}
\label{CtrParam}
\end{table*}

To generate pose references for each of the controllers, we perform an optimization to obtain a configuration satisfying the end-effector reference. Here, we use a one-step variant of \gls{AICO} \cite{toussaint2009robot}. 
Note, while the end-effector pose reference can be passed in directly to the end-effector controller, a postural optimization is used in this case to obtain a pose reference for the null-space or additional superimposed controllers. We extract the reference pose for each of the controllers using forward kinematics.

\subsection{Reference Trajectories}
The figure-of-8 (i.e., lemniscate) trajectory has been selected to show the dynamic behavior of the robot. The trajectory is composed of two orthogonal sinusoidal trajectories. The vertical trajectory has an amplitude of \SI{0.2}{\meter} and the transverse trajectory amplitude is \SI{0.1}{\meter}. The figure-of-8 trajectory is particularly demanding due to its multiple velocity inversions and wide joint movements range. Thus, introducing high variability of both the Jacobian and the inertial behavior of the robot.
We test the figure-of-8 reference motion in both simulation and hardware experiments.

In hardware experiments, we further test a sinusoidal trajectory with an amplitude of \SI{0.5}{\meter} and velocities up to about \SI{0.7}{\meter\per\second}. The straight-line experiment enabled us to test interaction and robustness at higher speeds.

\subsection{Simulation Experiments}
We simulate the robot using the Gazebo physics simulator and apply the Superimposition of Passive Task-Space Controllers control scheme directly without compensating for gravity, Coriolis, or other dynamic effects (in contrast to \cite{babarahmati2019}), i.e., as a model-free compliant controller.
For the simulation experiments, we compare nominal tracking performance with an interaction scenario where an unsensed environment obstacle has been introduced, cf. \autoref{fig:gazebo_unsensed_force_interaction}.

\subsection{Hardware Experiments}
In our hardware experiments, we use a Kuka LWR3+ robot. We control the manipulator using the \gls{FRI} at \SI{333.3}{\hertz} in \emph{joint impedance} mode with all gains set to zero to enable feed-forward torque control. Note, unlike our simulation experiments the Kuka's built-in controller compensates for dynamic effects and gravity.
On the real robot the tracking of the figure-of-8 trajectory has also been tested with and without a human operator applying random perturbations. The values used in the controller for the simulation and the experiments are reported in \autoref{CtrParam}. It shall also be noted that during the experiment we have kept the minimum set of controlled \gls{DoF} required to fully control the 3 \gls{DoF} of redundancy for the assigned tasks, being the task invariant to the configuration of the $7^{th}$ \gls{DoF} due to the symmetric geometry of the end-effector in the manipulator (\autoref{fig:intro}).

\section{Results}
\label{sec:results}

\begin{figure*}[t]
    \begin{subfigure}[t]{0.32\linewidth}
        \includegraphics[width=\linewidth,trim={0.1in 0.1in 0.12in 0.1in},clip]{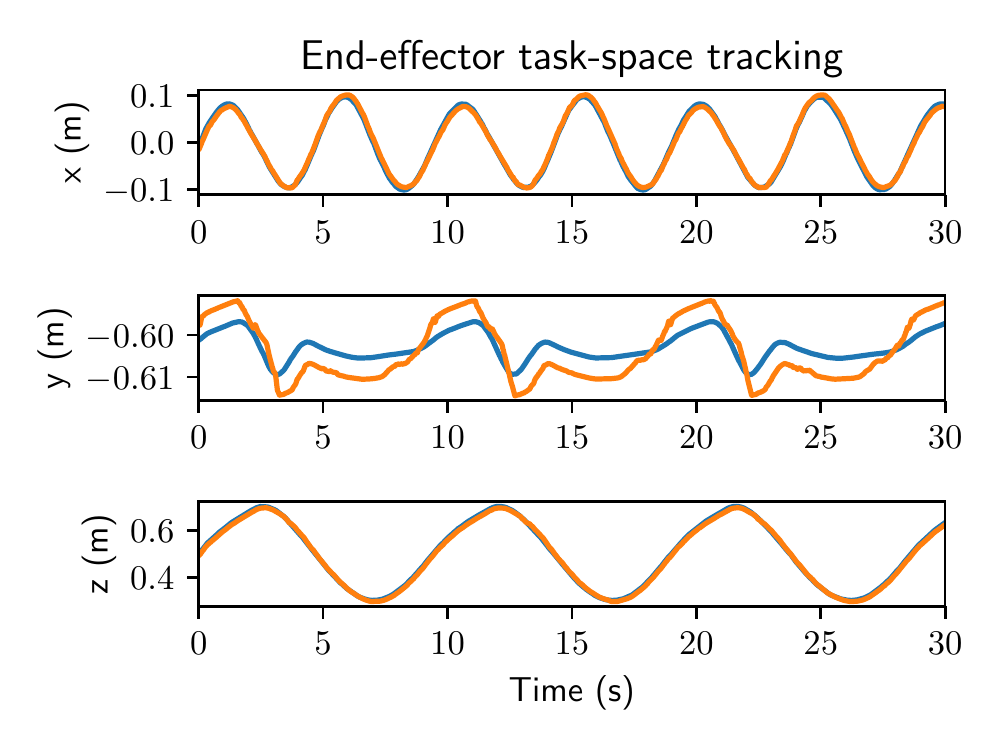}
        \caption{End-effector task-space trajectories showing reference and executed motion}
    \end{subfigure}
    \begin{subfigure}[t]{0.32\linewidth}
        \includegraphics[width=\linewidth,trim={0.1in 0.1in 0.12in 0.1in},clip]{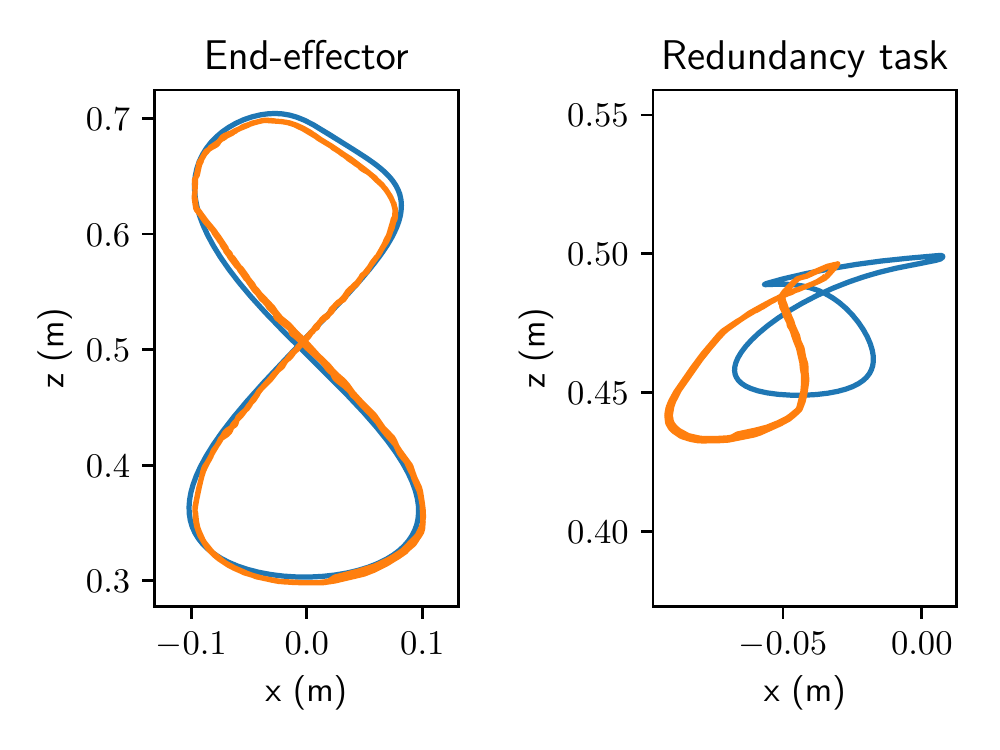}
        \caption{Task-space motion in $xz$-plane (reference in blue, executed in orange)}
    \end{subfigure}
    \begin{subfigure}[t]{0.32\linewidth}
        \includegraphics[width=\linewidth,trim={0.1in 0.1in 0.12in 0.1in},clip]{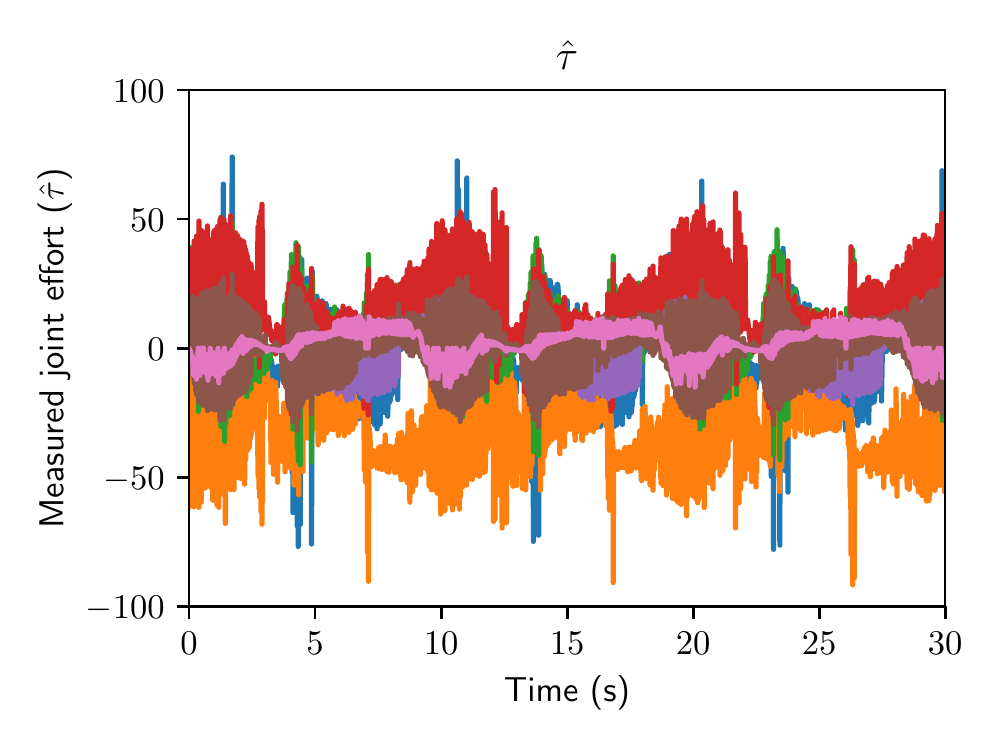}
        \caption{Measured joint space torques.\\}
    \end{subfigure}
    \caption{\emph{Simulation:} Fast figure-8, no interaction. 
    Note, as the simulation experiments are fully model-free (i.e., without gravity or non-linear effects compensation), the plotted torques are higher compared with \autoref{8-NOInt} as the \gls{FIC} automatically compensates for gravity and dynamic effects. This demonstrates that our controller can handle gravity/non-linear effects compensation as external disturbances $F_{ext}$ without the need for a dynamics model.}
    \label{fig:Sim_8-NOInt}
\end{figure*}


\begin{figure*}[t]
    \begin{subfigure}[t]{0.32\linewidth}
        \includegraphics[width=\linewidth,trim={0.1in 0.1in 0.12in 0.1in},clip]{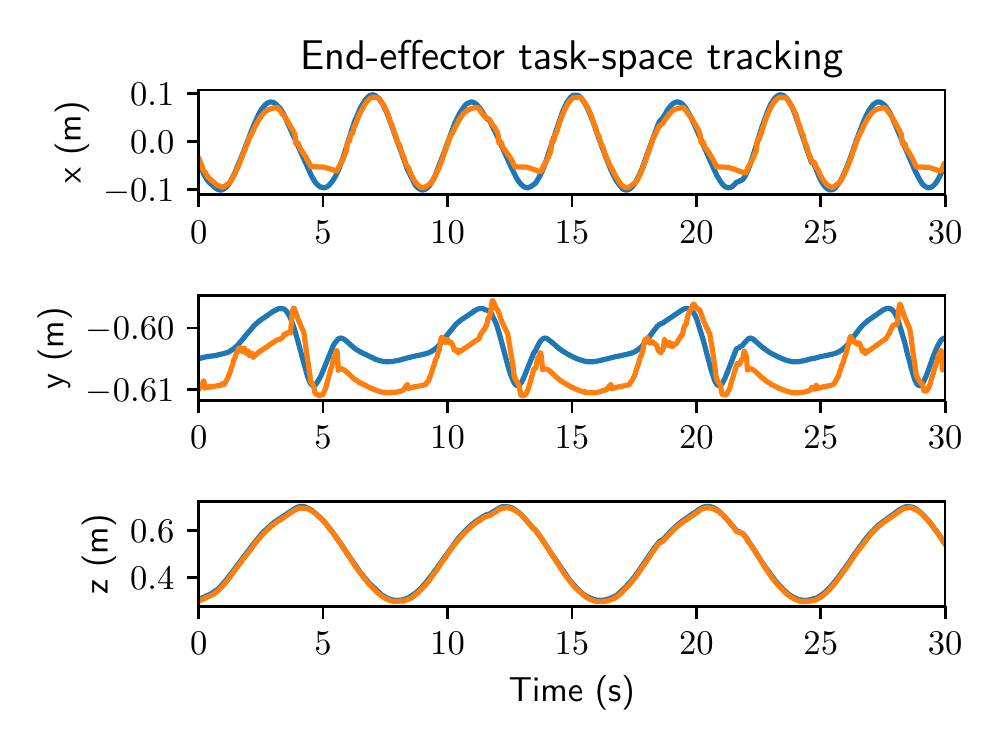}
        \caption{End-effector task-space trajectories showing reference and executed motion}
    \end{subfigure}
    \begin{subfigure}[t]{0.32\linewidth}
        \includegraphics[width=\linewidth,trim={0.1in 0.1in 0.12in 0.1in},clip]{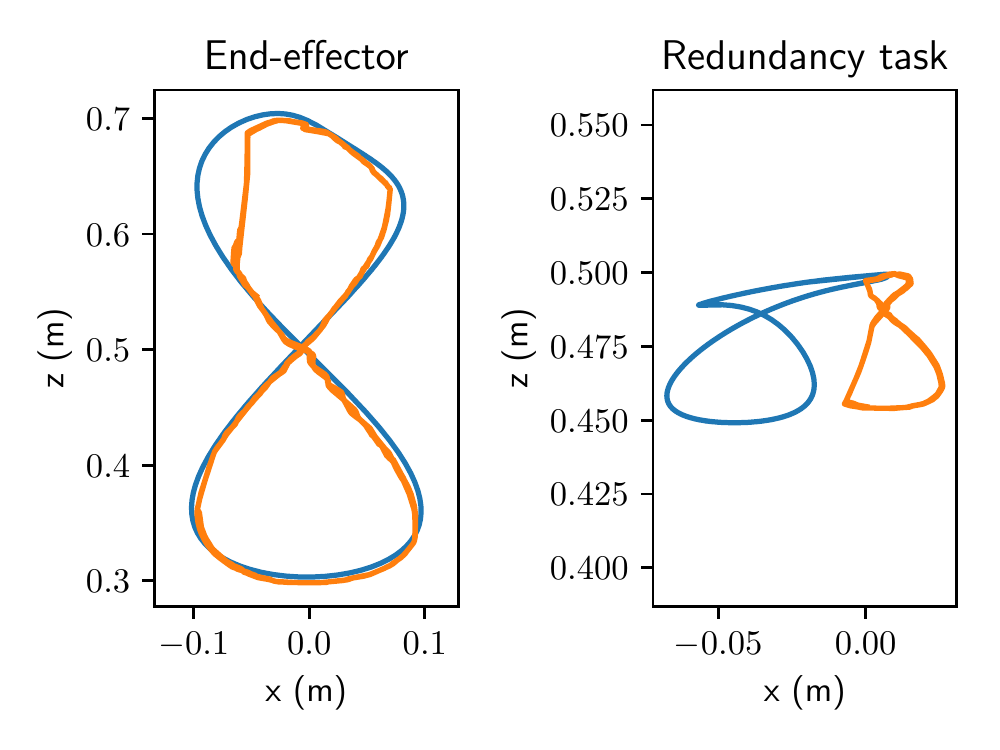}
        \caption{Task-space motion in $xz$-plane (reference in blue, executed in orange)}
    \end{subfigure}
    \begin{subfigure}[t]{0.32\linewidth}
        \includegraphics[width=\linewidth,trim={0.1in 0.1in 0.12in 0.1in},clip]{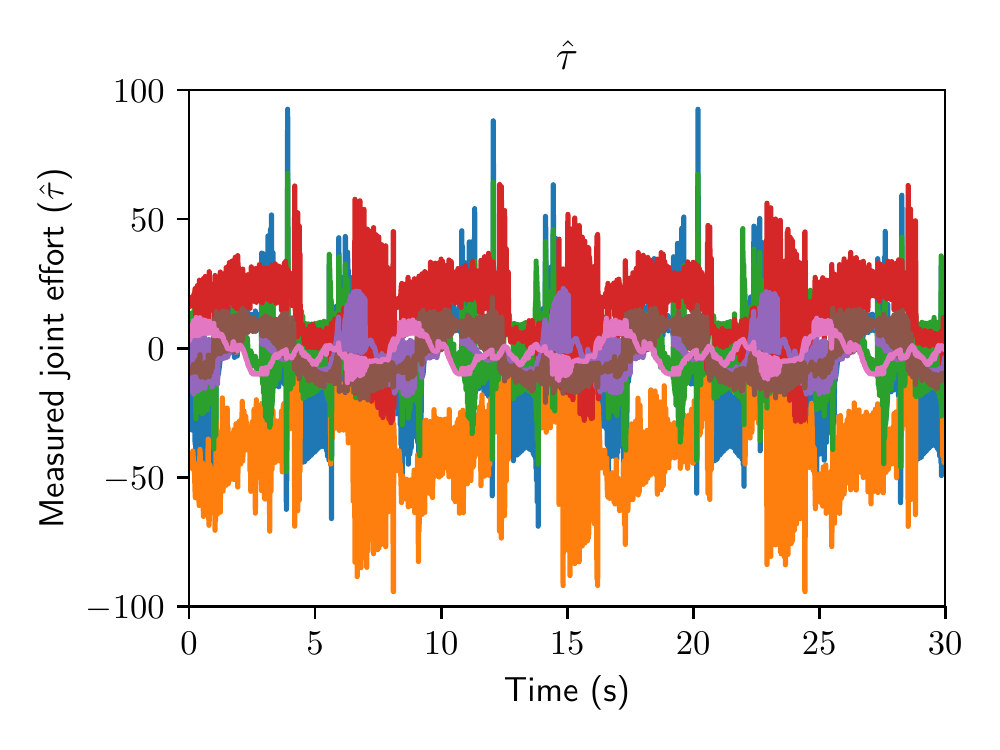}
        \caption{Measured joint space torques.\\}
    \end{subfigure}
    \caption{\emph{Simulation:} Fast figure-8, interaction with unsensed obstacle (cf. \autoref{fig:gazebo_unsensed_force_interaction}). The stack of passive task-space controllers compensates for the obstacle using the redundancy task by smoothly sliding across its surface, cf. (b). Note, that despite the interaction there are no torque spikes or instability, cf. (c).}
    \label{fig:Sim_8-WInt}
\end{figure*}
\begin{figure*}[t]
    \begin{subfigure}[t]{0.32\linewidth}
        \includegraphics[width=\linewidth,trim={0.1in 0.1in 0.1in 0.1in},clip]{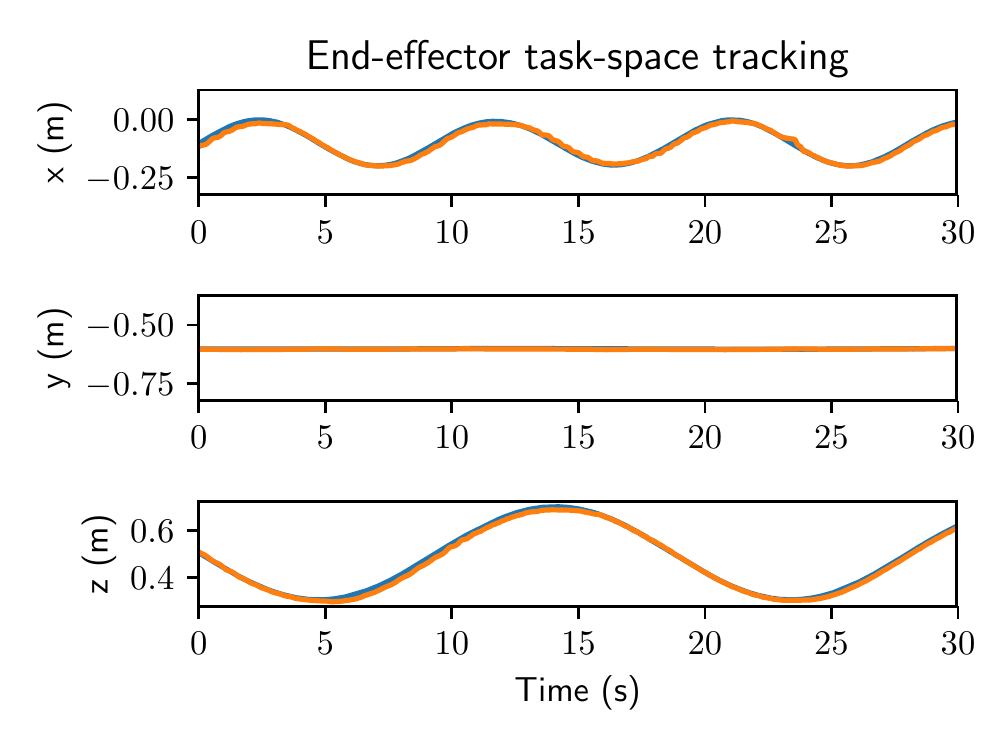}
        \caption{End-effector task-space trajectories showing reference and executed motion}
    \end{subfigure}
    \begin{subfigure}[t]{0.32\linewidth}
        \includegraphics[width=\linewidth,trim={0.1in 0.1in 0.12in 0.1in},clip]{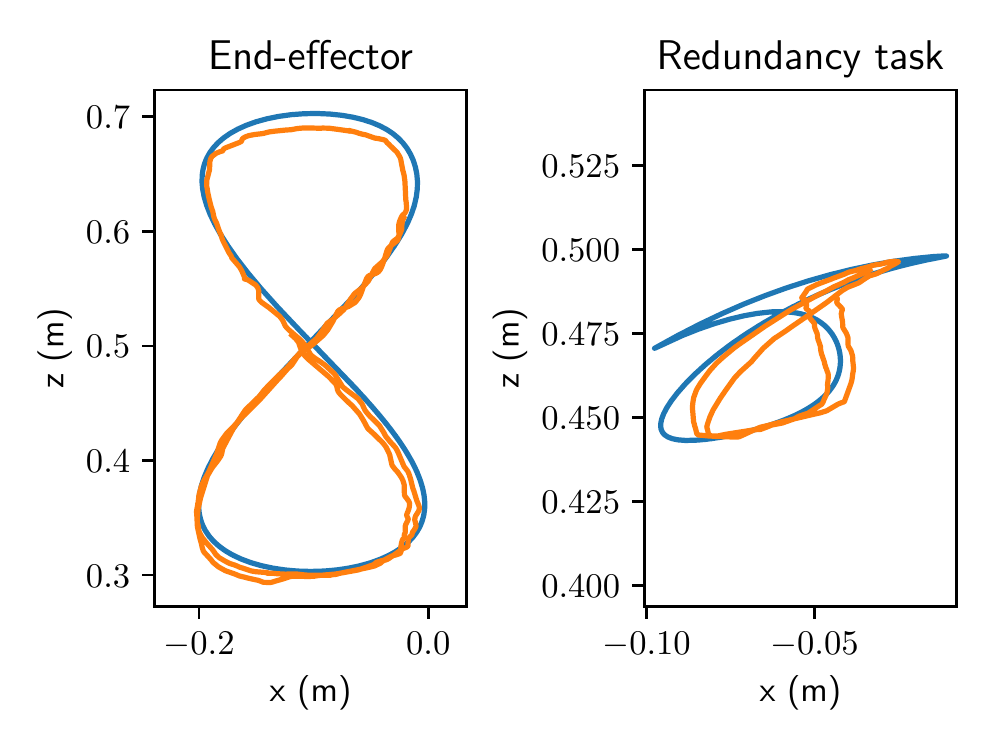}
        \caption{Task-space motion in $xz$-plane (reference in blue, executed in orange)}
    \end{subfigure}
    \begin{subfigure}[t]{0.32\linewidth}
        \includegraphics[width=\linewidth,trim={0.1in 0.1in 0.10in 0.1in},clip]{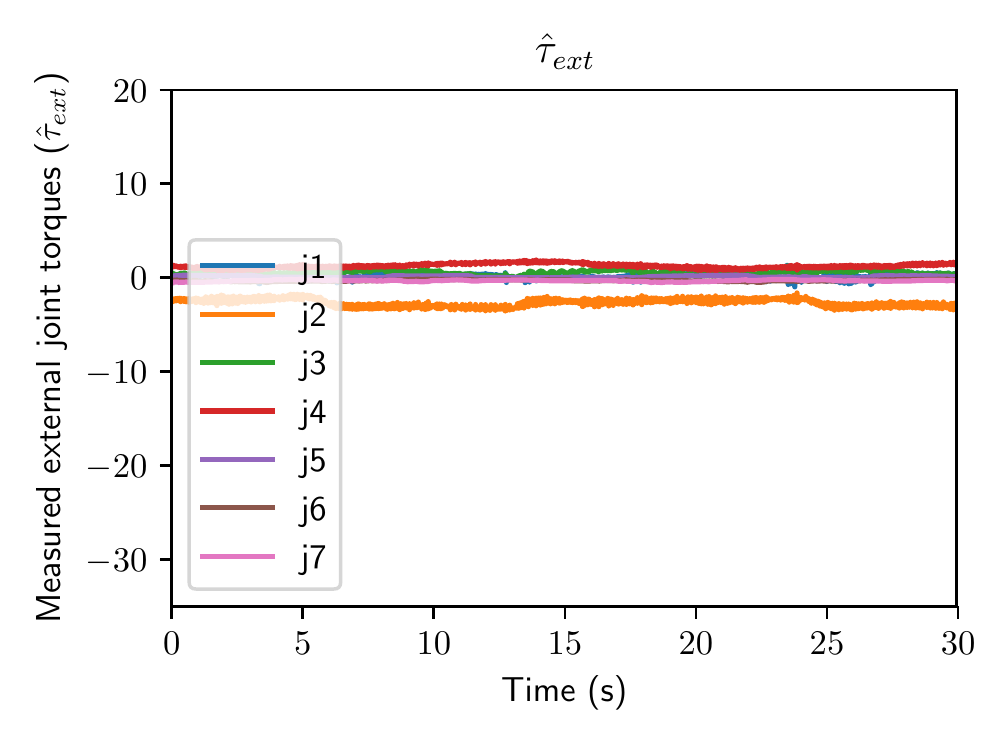}
        \caption{Measured external torques (interaction forces) at the joints.}
    \end{subfigure}
    \caption{\emph{Hardware experiments:} Fast figure-8, no interaction. The controller tracks the reference closely both for the primary and redundancy tasks.} 
    \label{8-NOInt}
\end{figure*}

\begin{figure*}[t]
    \begin{subfigure}[t]{0.32\linewidth}
        \includegraphics[width=\linewidth,trim={0.1in 0.1in 0.12in 0.1in},clip]{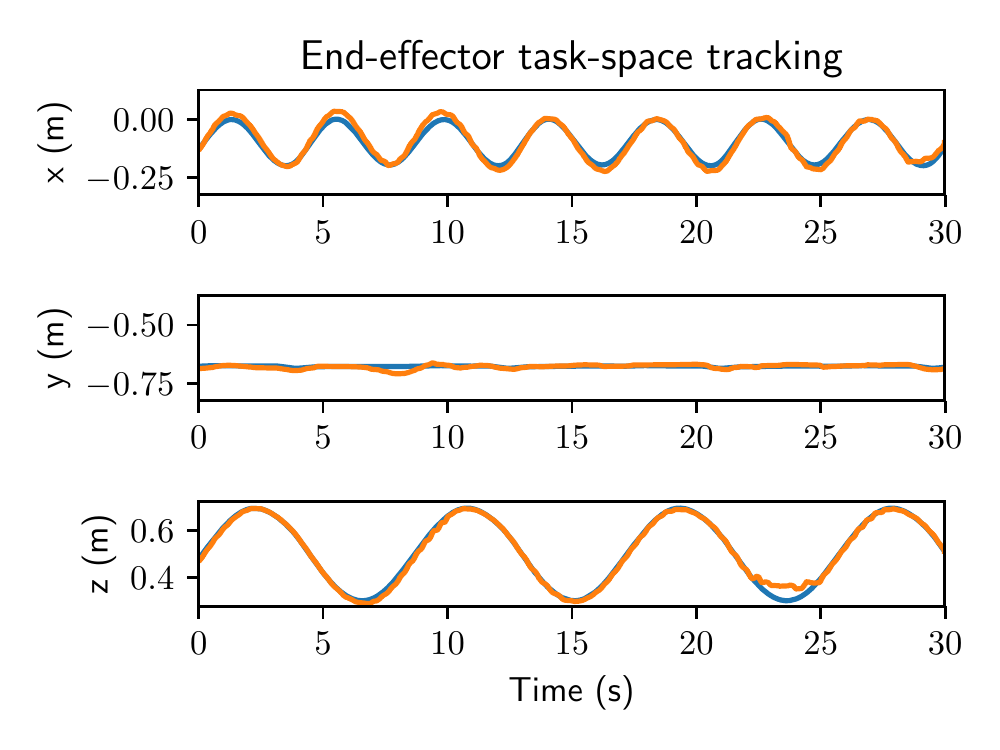}
        \caption{End-effector task-space trajectories showing reference and executed motion}
    \end{subfigure}
    \begin{subfigure}[t]{0.32\linewidth}
        \includegraphics[width=\linewidth,trim={0.1in 0.1in 0.12in 0.1in},clip]{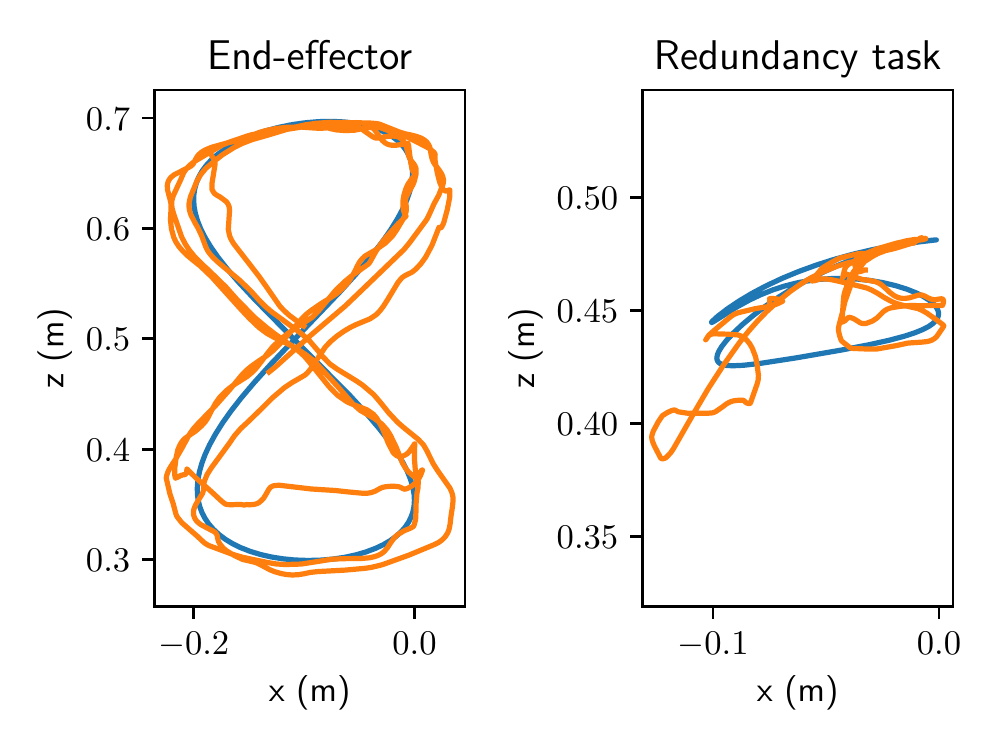}
        \caption{Task-space motion in $xz$-plane (reference in blue, executed in orange)}
    \end{subfigure}
    \begin{subfigure}[t]{0.32\linewidth}
        \includegraphics[width=\linewidth,trim={0.1in 0.1in 0.12in 0.1in},clip]{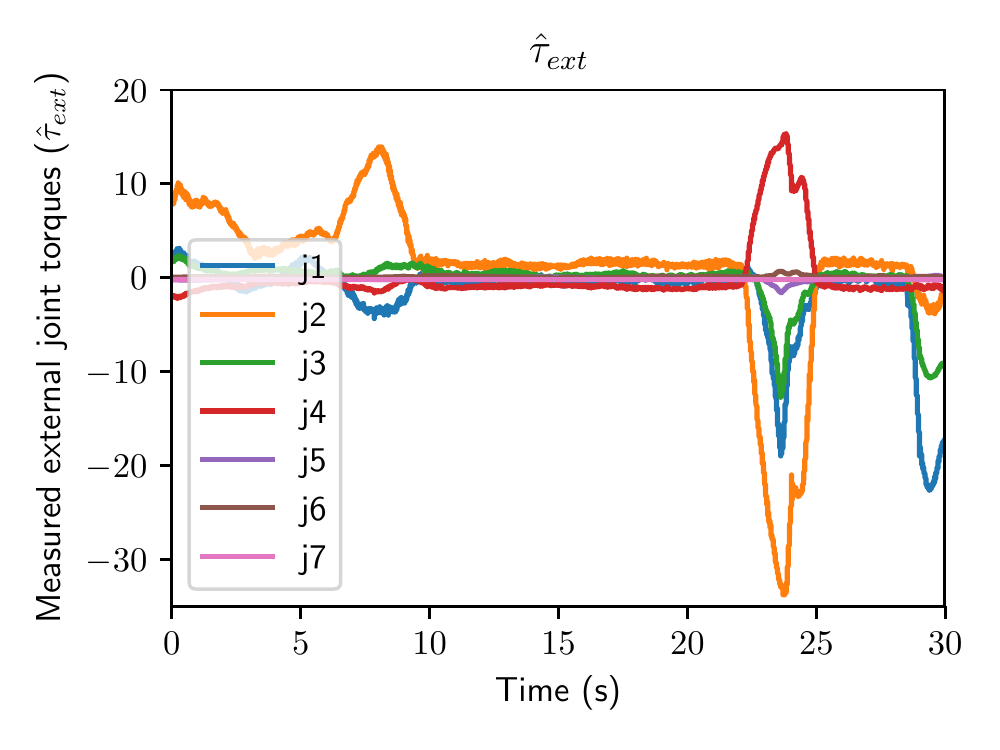}
        \caption{Measured external torques (interaction forces) at the joints.}
    \end{subfigure}
    \caption{\emph{Hardware experiments:} Fast figure-8, with interaction. Note, the significant external joint torque forces compared with \autoref{8-NOInt} due to the interaction/perturbations.}
    \label{8-Int}
\end{figure*}

To complement the plots in this section, the reader is recommended to watch the supplementary video demonstrating the tracking and interaction both in simulation and hardware experiments. We also include a sequence demonstrating the safe behavior of the controller during calibration of the \gls{FIC} parameters given in \autoref{CtrParam}.

The simulation results are shown in \autoref{fig:Sim_8-NOInt} for the free motion, and in \autoref{fig:Sim_8-WInt} for the interaction behavior. They show that the robot can be successfully controlled without dynamic compensation, and that it can achieve dexterous dynamic behaviors.
The tracking \glspl{RMSE} at the end-effector are recorded without interaction as RMSE$_\text{x}=\SI{5.6}{\milli\meter}$,  RMSE$_\text{y}=\SI{4.6}{\milli\meter}$, and  RMSE$_\text{z}=\SI{6.1}{\milli\meter}$. For simulation with interaction with an obstacle: RMSE$_\text{x}=\SI{13.2}{\milli\meter}$,  RMSE$_\text{y}=\SI{4.2}{\milli\meter}$, and  RMSE$_\text{z}=\SI{5.5}{\milli\meter}$.
I.e., the tracking performance degrades in one dimension impacted by the obstacle ($x$), while being virtually unaffected in $y$ and $z$.

The experimental data for the hardware experiments of the figure-of-8 trajectory are shown in \autoref{8-NOInt} and \autoref{8-Int}.
The errors recorded during free motion are: RMSE$_\text{x}=7.6$ \si{\milli\meter}, RMSE$_\text{y}=1.5$ \si{\milli\meter}, and RMSE$_\text{z}=8.6$ \si{\milli\meter}.
The perturbations do not affect the tracking performance at the end-effector task, but they are fully compensated by the deflection from the secondary task target at the elbow joint, as shown in \autoref{8-Int}b. 
The \gls{RMSE} during interaction are RMSE$_\text{x}=20.3$ \si{\milli\meter}, RMSE$_\text{y}=7.9$ \si{\milli\meter}, and RMSE$_\text{z}=9.1$ \si{\milli \meter}.

\begin{figure*}[t]
    \begin{subfigure}[t]{0.32\linewidth}
        \includegraphics[width=\linewidth,trim={0.1in 0.1in 0.1in 0.1in},clip]{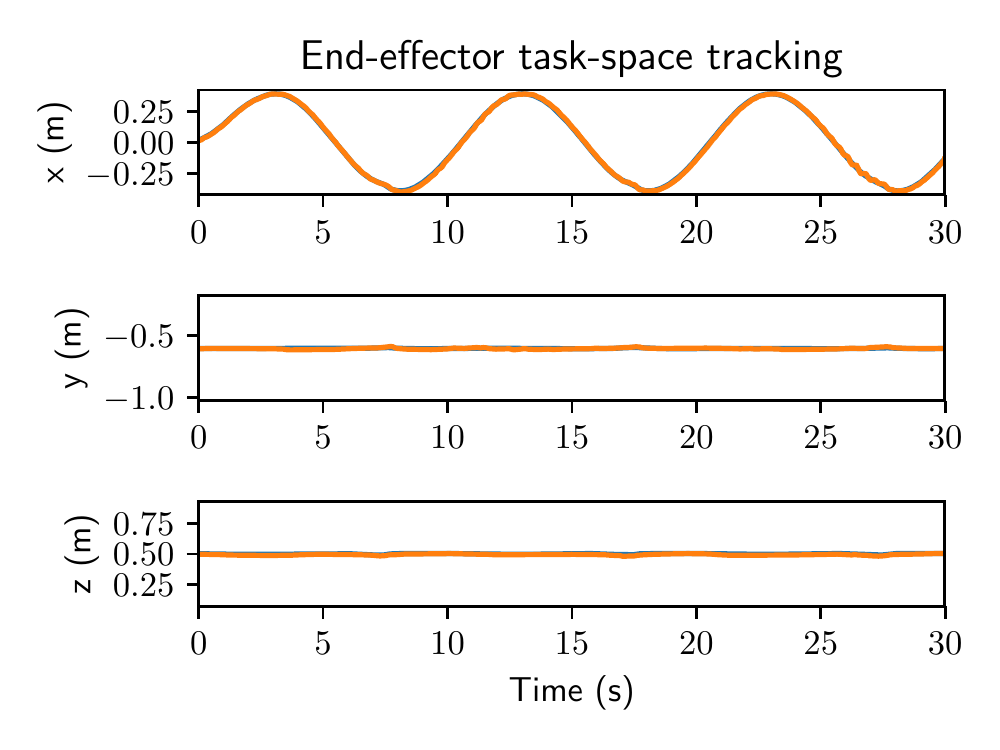}
        \caption{End-effector task-space trajectories showing reference and executed motion}
    \end{subfigure}
    \begin{subfigure}[t]{0.32\linewidth}
        \includegraphics[width=\linewidth,trim={0.1in 0.1in 0.12in 0.1in},clip]{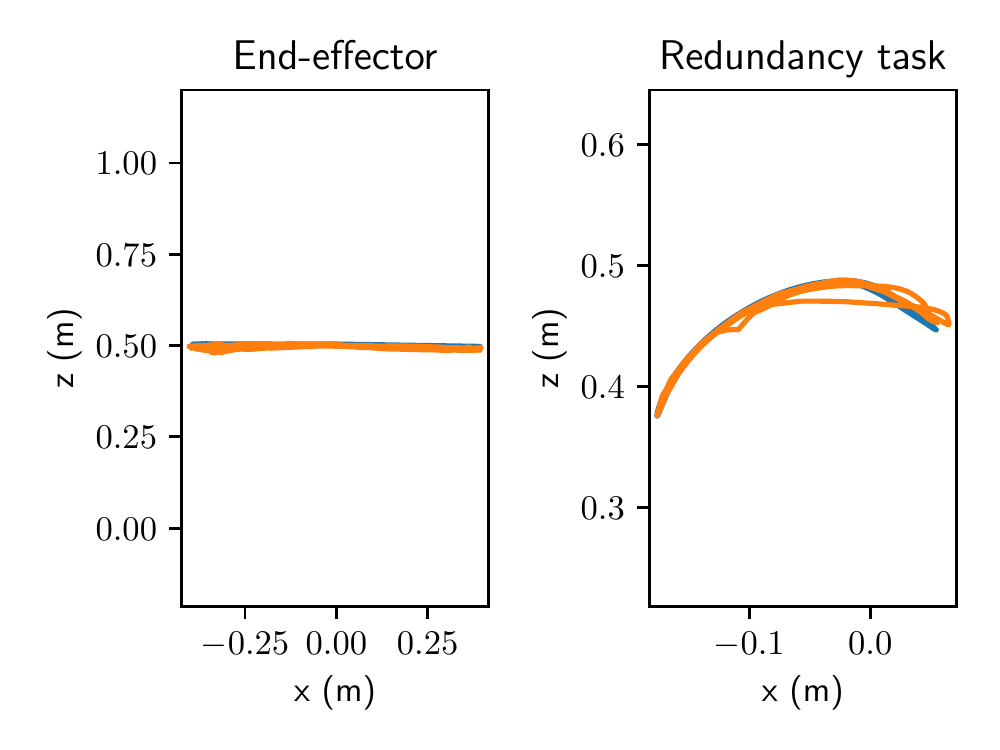}
        \caption{Task-space motion in $xz$-plane (reference in blue, executed in orange)}
    \end{subfigure}
    \begin{subfigure}[t]{0.32\linewidth}
        \includegraphics[width=\linewidth,trim={0.1in 0.1in 0.12in 0.1in},clip]{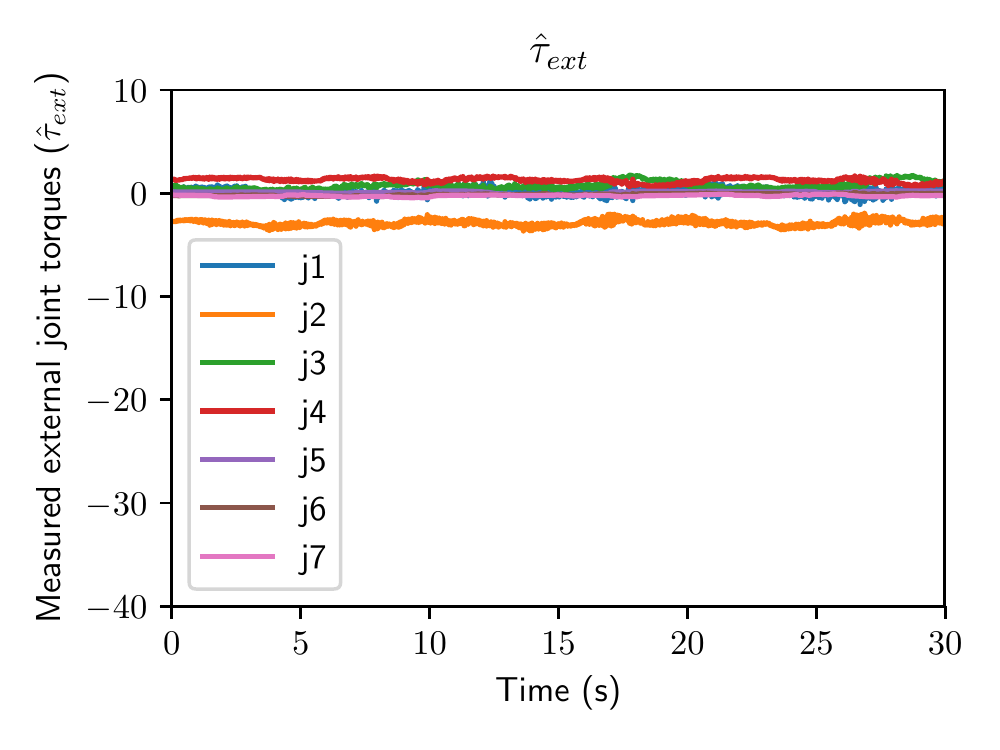}
        \caption{Measured external torques (interaction forces) at the joints.}
    \end{subfigure}
    \caption{\emph{Hardware experiments:} \emph{Straight line reference, no interaction:} The robot tracks the task-space references for both the primary and redundancy task closely
    joint configurations are sacrificed.}
    \label{Line-NoInt}
\end{figure*}

The results for the straight-line trajectory experiment (\autoref{Line-NoInt} and \autoref{line-Int}) show an ability of the controller to complete the task and reject perturbations by reducing tracking on the secondary task.
The errors are RMSE$_\text{x}=6.1$ \si{\milli\meter}, RMSE$_\text{y}=4.6$ \si{\milli\meter}, and RMSE$_\text{z}=6.7$ \si{\milli \meter}. 
The RMSE for interaction are RMSE$_\text{x}=9.6$ \si{\milli\meter}, RMSE$_\text{y}=7.7$ \si{\milli\meter}, and RMSE$_\text{z}=7.3$ \si{\milli\meter}.

It shall also be remarked how the robot remained safe to interact with despite the high joint feed-forward torques involved in the motions, which reached $\approx30$ \si{\newton\meter} for both the 8 trajectory and the linear trajectory.

\section{Discussion}
\label{sec:discussion}
The results show the proposed method enables an intrinsically stable control framework for redundant robots which does not rely on inverse dynamics and projection matrices.
The proposed method is robust to unknown environmental interactions and singularities, where safe means that the robot does not show erratic behaviors even while perturbed or when there is a sudden change in the desired task (e.g., sudden acceleration/deceleration). The RMSE data show how the robot keeps the minimum tracking accuracy (i.e., maximum error) contained under \SI{1}{\centi\meter} for the unperturbed experiments, which is in line with the task requirement set in the controller parameters $x_\text{b}=\SI{1.1}{\centi\meter}$. This results are in line with the results obtained in \cite{babarahmati2019}, and they are lower than other impedance controller frameworks that usually have error of few centimeters \cite{dietrich2015,Dietrich2016}.

The introduction of significant perturbation degrades the minimum tracking accuracy to \SI{2}{\centi\meter}, but the controller remains stable and it is able to recover once the perturbation ends. It shall be remembered that the trade-off between accuracy and robustness is a known trade-off in interaction control frameworks \cite{ott2010unified}.  While admittance control may provide better accuracy, it requires accurate knowledge of interaction force intensity and direction in all the points of contact with the environment. On the other hand, impedance controllers provide better safety of interaction but sacrifice tracking accuracy in favor of compliance \cite{ott2010unified,xin2020}. Variable impedance controllers have been proposed as a solution to this dilemma, but the stability requirements on the impedance updates are often very stringent and difficult to retain under highly variable environmental conditions \cite{angelini2019online,li2018force}. The proposed framework provides the better of both worlds providing good tracking accuracy while retaining the robustness typical of impedance controllers; furthermore, it enables online adjustment of the impedance profiles \cite{babarahmati2019}.

The data also confirm our hypothesis that redundant robot interaction behavior can be accurately defined without any \textit{a priori} knowledge of the system dynamics model, being \autoref{TControllerStack} the control command. 
In our simulation experiments, we show that the tracking performance in this work is similar to the results reported in \cite{babarahmati2019} that relied on a compensation of the robot dynamics and the use of a null-space controller. This latest result is particularly important because robots' mechanical properties such as inertia and joints friction matrices are often difficult to retrieve and highly unreliable \cite{bruder2019, vasudevan2017}.
The knowledge of the actuation characteristics and the kinematic structure are still necessary for the implementation of the proposed method. However, they are both normally accurate, and easier to obtain if not available.   

\begin{figure*}[t]
    \begin{subfigure}[t]{0.32\linewidth}
        \includegraphics[width=\linewidth,trim={0.1in 0.1in 0.1in 0.1in},clip]{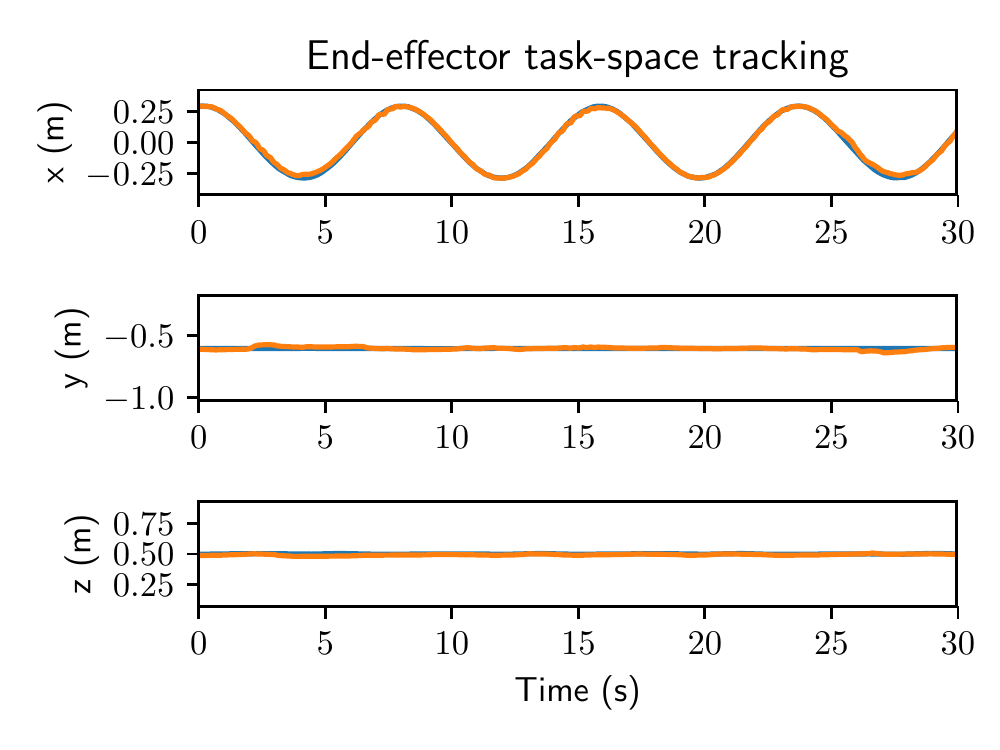}
        \caption{End-effector task-space trajectories showing reference and executed motion}
    \end{subfigure}
    \begin{subfigure}[t]{0.32\linewidth}
        \includegraphics[width=\linewidth,trim={0.1in 0.1in 0.12in 0.1in},clip]{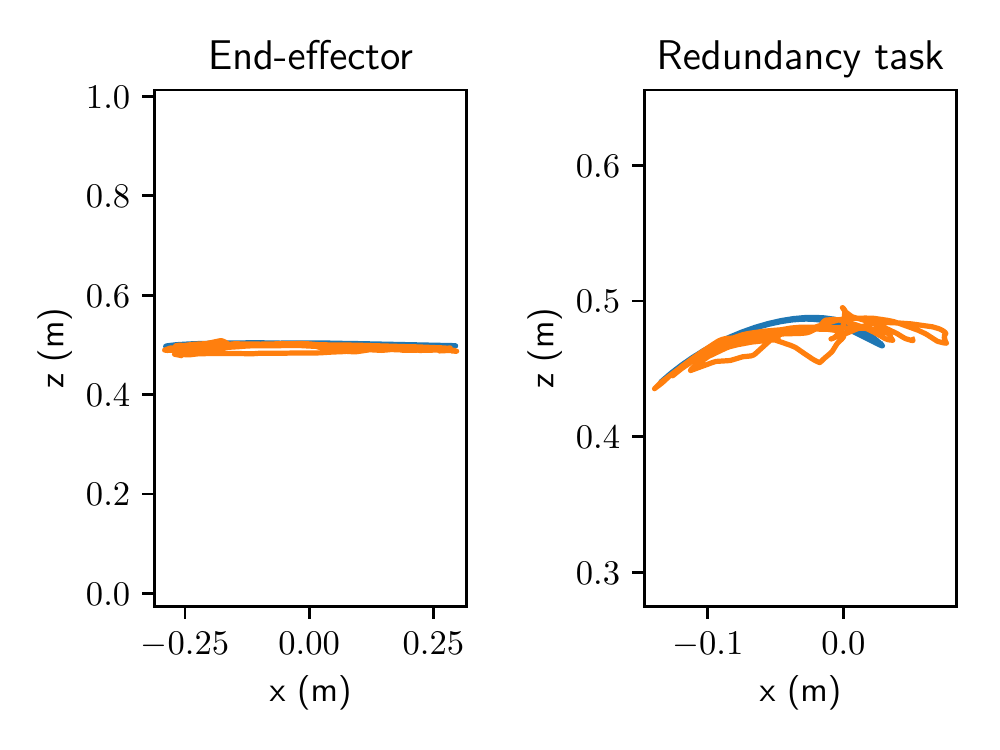}
        \caption{Task-space motion in $xz$-plane (reference in blue, executed in orange)}
    \end{subfigure}
    \begin{subfigure}[t]{0.32\linewidth}
        \includegraphics[width=\linewidth,trim={0.1in 0.1in 0.1in 0.1in},clip]{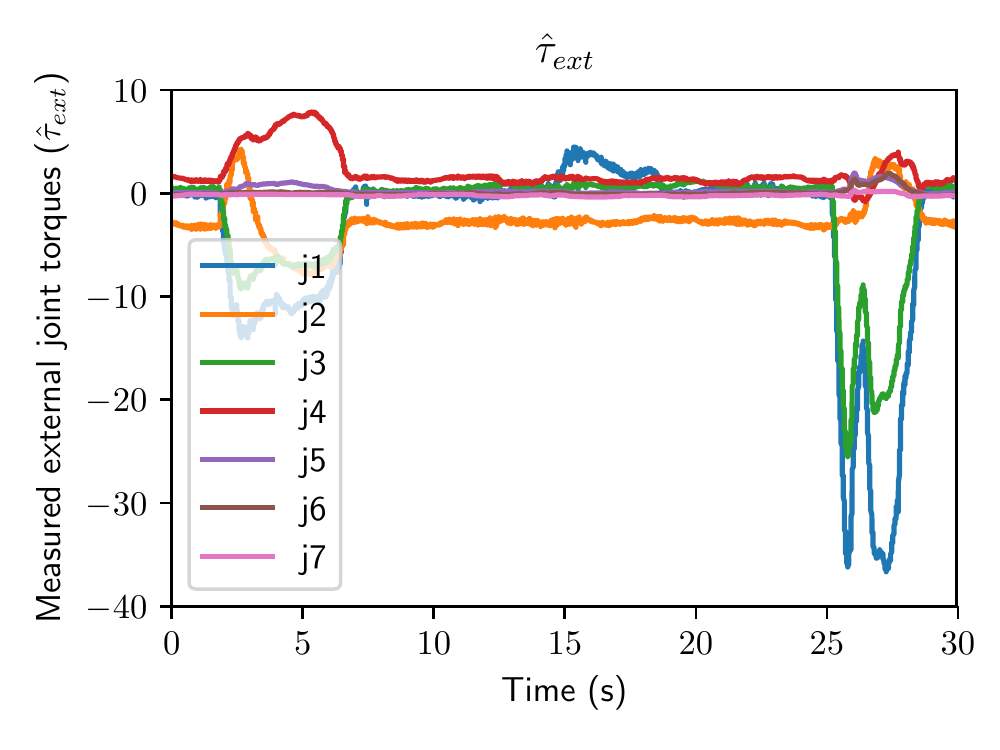}
        \caption{Measured external torques (interaction forces) at the joints.}
    \end{subfigure}
    \caption{\emph{Hardware experiments:} \emph{Straight line reference (\SIrange{0.4}{0.7}{\metre\per\second}), with interaction:} Robot joint trajectories and effort during interaction experiments. The robot joints diverged significantly from the planned configuration, and reached position across singularities without any evident impact on the robot performance in terms of robustness and stability.
    }
    \label{line-Int}
\end{figure*}

The FIC controller generates an asymptotically stable potential field around the target state that enables direct superimposition of multiple controllers without compromising the system stability. 
The controller superimposition generates a force field that acts as a trade-off cost function determining the preferred path of motion in robot's configuration space. 
The force upper-bound of the controllers guarantees that the loss of accuracy in the main task is contained $x_\text{b}$ until the condition are compatible with the mechanical characteristics of the system. 
Especially, if we consider that the proposed method is a compliant postural controller, where accurate tracking is subordinate to robustness of interaction. 
In other words, the controller stabilizes the robot around a desired posture relying on the non-linear stiffness profile to compensate for its non-linear and environmental interaction, sacrificing the redundancy task before degrading the end-effector task beyond the selected accuracy. 
This is confirmed by both the simulation and the experimental data, showing how the FIC controller tries and successfully keeps the accuracy under \SI{0.11}{\centi\meter} in unperturbed conditions. 
The data also describe that the controller fully sacrifices the redundancy task in the attempt of retaining the same accuracy while experiencing external perturbation that exceed its mechanical limits, as shown in \autoref{fig:Sim_8-WInt}, \autoref{8-Int} and \autoref{line-Int}.

The data show that the proposed method can achieve a highly dynamic interaction using variable impedance at the controller level. The FIC also enables online tuning of the impedance behavior and is robust to reduce bandwidth in the feedback signals \cite{babarahmati2019}, allowing to switch from rigid to soft behaviors seemingly. Nevertheless, the performances are strictly related to the physical hardware capabilities, and a higher band-pass in the mechanics of the robot implies a higher stiffness to mass ratio. Therefore, it will be interesting to study these capabilities by deploying in hardware equipped with VSA to conduct a systematic experiment on these properties. Furthermore, it may also enable the switching from a model-based (e.g., the one proposed in \cite{keppler2018}) to a data-driven control of their non-linear actuators dynamics. It is worth also noting that, at the current stage, it is impossible to compare the results of these types of architectures due to their different hardware requirements.
Nevertheless, we can say that both of them achieve non-linear impedance behavior and robustness to highly dynamic interactions. The fractal impedance controller hardware requirements are less stringent, and it has a simpler formulation. The results presented in \cite{keppler2018} indicate that the DLR robot can achieve a stiffer behavior. However, it is impossible to discriminate if they are connected solely to a hardware superiority or there is also a controller component in play.

The superimposition of task-space controllers also opens new possibilities for improving controllability and dexterity for compliant robots, developing human motor control theory, and robust control architecture for learning algorithms. 
In fact, the dynamics of soft robots are even more challenging to model than rigid dynamics \cite{bruder2019}, as the dynamic modelling of robots is founded on the assumption of rigidity \cite{siciliano}. In regards to human motor control, having a framework that enables robustness and dexterity of interaction will enable to overcome the current limitation of the \gls{PMP} model, which still relies on inverse matrices for trajectory optimization \cite{tommasino2017,tiseo2018}. 
Finally, the learning algorithms are currently facing the challenges of performing a system identification to guarantee the stability of the learned behavior. 
The proposed method removes this challenge and the learning component can focus on learning how to synchronize the task-space controller to maximize the efficiency and the dexterity of the robot.

\section{Conclusions} \label{sec:conclusion}
The experimental results confirmed our hypothesis that it is possible to control a redundant robot with a superimposition of task-space controllers.
This approach renders the architecture intrinsically robust to singularity and fully passive which guarantees stability. 
It is important to properly balance the strength of the controllers to guarantee that the secondary tasks do not interfere with the end-effector controller, which may result challenging under certain conditions. Nevertheless, unbalanced controllers may interfere with the action efficacy, but not with the robustness and stability of interaction. 

The proposed framework does not require any \textit{a priori} knowledge of the system dynamics parameters (i.e., Inertia, Friction, and Gravity).
It suited for applications where the stability of interaction to unpredictable environments is more important than the tracking accuracy. 
Future work will focus on improving the coordination among the secondary task-space controllers to improve the tracking accuracy of the end-effector task as well as systems were coupling effects may be introduced.
\appendix
\section{Lyapunov Stability Proof}
\label{sec:stabilityAnalysis}
The fractal attractor generates a piecewise-smooth energy manifold around the desired state, shown in \autoref{fig:2b}. The autonomous trajectory of the attractor have an energy flow that is always converging to the null state $(\tilde{x}=0,~\dot{x}=0)$.  Furthermore, the intrinsic damping can be assumed to be zero without loss of generality in the proposed controller being the desired velocity always zero (i.e., $\dot{x}_\text{d}:=0$). 
Let us now consider the controller's autonomous dynamics for a mono-dimensional mass-spring system:
\begin{equation}
	\label{Eq:dynamicdot}
	   \left\{
	   \begin{array}{ll}
	    \Lambda(x)\ddot{x} +\dot{\Lambda}(x)\dot{x}  + K_\text{d}(\tilde{x})\tilde{x}=0 & \text{ Divergence}\\
	    \Lambda(x)\ddot{x} +\dot{\Lambda}(x)\dot{x} + K_{\text{c}}(\tilde{x}_\text{max})\tilde{x}=0 & \text{ Convergence}
	    \end{array}\right.
\end{equation}
	
\noindent {where $\Lambda (x)$ is the task-space inertia. $\dot{\Lambda}(x)\dot{x}$ are the Coriolis and Centrifugal forces, and $\tilde{x}_\text{max}$ is the maximum displacement reached during the divergence phase. A valid Lyapunov's candidate is:}
\begin{equation}
	\label{eq:LyapunvFunction}
	\begin{cases}
        \displaystyle{	V_\text{d}= \frac{\dot{x}^{T}\Lambda(x) \dot{x}}{2} +\int K_{d}(\tilde{x})\tilde{x}~d\tilde{x}} \\ 
	    \displaystyle{V_\text{c}= \frac{\dot{x}^T\Lambda (x)\dot{x}}{2} +\frac{\tilde{x}^T K_\text{c}(\tilde{x}_\text{max})\tilde{x}}{2}+\frac{\int_{0}^{\tilde{x}_{\text{max}}} K_{d}(\tilde{x})\tilde{x}~d\tilde{x}}{2}} 
	\end{cases}
\end{equation}
\noindent {V time derivative is:}
	\begin{equation}
	\label{Eq:Vdot}
	   \begin{cases}
	        \dot{V}_\text{d}=(\Lambda(x)\ddot{x} +\dot{\Lambda}(x)\dot{x}  + K_d(\tilde{x})\tilde{x})\dot{x}=0\\
	        \dot{V}_\text{c}=(\Lambda(x)\ddot{x} +\dot{\Lambda}(x)\dot{x} + K_{\text{c}}(\tilde{x}_\text{max})\tilde{x})\dot{x}=0
	    \end{cases}
	\end{equation}
  
 \autoref{eq:LyapunvFunction} and \autoref{Eq:Vdot} prove the stability of the two pieces that compose the attractor autonomous trajectories. However, Lyapunov's stability for non-smooth systems also requires to verify that the candidate is a Lipschitz function. In other words, $V$ is continuous and $\dot{V}$ is finite in the transition points. The condition is imposed when switching from divergence to convergence ($\tilde{x}\ne0$) by the conservation of energy principle, which is applied to derive \autoref{alg:FIC}. The validity of this condition in $\tilde{x}=0$ can be verified as follows: 
 
	\begin{equation}
	\label{eq:LyapunvFunction2}
	\begin{cases}
	\displaystyle{\lim_{\substack{\tilde{x}\to0 \\ \dot{x}\to0}}{V_\text{d}}=\int K_{d}(\tilde{x})\tilde{x}~d\tilde{x}=0}\\
    \displaystyle{\lim_{\substack{\tilde{x}\to0 \\ \dot{x}\to0}}{V_\text{c}}=\frac{\tilde{x}^T K_\text{c}(\tilde{x}_\text{max})\tilde{x}}{2}+0.5\int_{0}^{\tilde{x}_\text{max}} K_\text{d}(\tilde{x})\tilde{x}~d\tilde{x}=}
    \\~~~~~~=\displaystyle{-~0.5\int_{0}^{\tilde{x}_\text{max}} K_\text{d}(\tilde{x})\tilde{x}~d\tilde{x}}~+\\
    ~~~~~~~\displaystyle{+~0.5\int_{0}^{\tilde{x}_{\text{max}}} K_\text{d}(\tilde{x})\tilde{x}~d\tilde{x}=0}
	\end{cases}
	\end{equation}
Furthermore, the $\dot{V}=0$ in both the transition states $(\tilde{x}_{\text{max}})$ and $(\dot{x}=0)$, which determines that the controller respects all the conditions required for stability.

\bibliography{references}

\end{document}